\documentclass[journal]{preamble/new-aiaa}

\usepackage[utf8]{inputenc}
\usepackage{textcomp}

\usepackage{graphicx}
\usepackage{amsmath}
\usepackage[version=4]{mhchem}
\usepackage{siunitx}
\usepackage{longtable,tabularx}
\usepackage{mathtools}
\usepackage{tikz}
\usepackage{amsmath}
\usepackage{amsfonts}
\usepackage{xcolor}
\usepackage{graphicx}
\graphicspath{{figures/}}
\usepackage{dsfont} 
\usepackage{cleveref}
\usepackage{svg}
\usepackage{preamble/vector} 
\usepackage{caption}
\usepackage{subcaption}
\usepackage{adjustbox}
\usepackage{longtable,tabularx,booktabs}
\usepackage[flushleft]{threeparttable}
\usepackage{soul}
\usepackage{multirow}
\usepackage{lipsum}
\usepackage{makecell}
\usepackage[nobreak=true]{mdframed}
\usepackage{comment}
\usepackage{afterpage}
\usepackage{setspace}
\usepackage{silence}
\definecolor{backgroundgray}{HTML}{F2F2F2}

\hypersetup{hidelinks}

\newenvironment{juliaframe}{%
    \linespread{1.25}\selectfont
    \begin{mdframed}[%
        backgroundcolor=backgroundgray,%
        hidealllines=true,%
        innerleftmargin=0pt,%
        innertopmargin=-1pt,%
        innerbottommargin=-6pt,%
        innerrightmargin=0pt,
        skipabove=12pt,
        skipbelow=-6pt,
    ]
}{%
    \end{mdframed}
}

\newsavebox\CBox

\definecolor{darkgreen}{HTML}{38761D}
\definecolor{darkred}{HTML}{980000}
\definecolor{gray1}{HTML}{999999}
\definecolor{gray2}{HTML}{CCCCCC}
\definecolor{gray3}{HTML}{D9D9D9}
\definecolor{gray4}{HTML}{EFEFEF}

\usetikzlibrary{calc}
\usetikzlibrary{shapes.geometric}
\usetikzlibrary{external}
\usetikzlibrary{patterns}
\usetikzlibrary{shapes,arrows,fit}
\usetikzlibrary{positioning}
\usetikzlibrary{arrows.meta, calc, shapes}
\usetikzlibrary{graphs}
\usetikzlibrary{decorations.pathmorphing}
\usetikzlibrary{decorations.pathreplacing}

\DeclareMathOperator*{\argmax}{arg\,max}

\renewcommand{\vec}[1]{\vect{#1}}

\usepackage{siunitx}
\renewcommand{\thefootnote}{\fnsymbol{footnote}}
\newcommand{\numericfootnote}[1]{%
\let\oldthefootnote=\thefootnote%
\stepcounter{mpfootnote}%
\addtocounter{footnote}{-1}%
\renewcommand{\thefootnote}{\arabic{mpfootnote}}%
\footnote{#1}%
\let\thefootnote=\oldthefootnote%
}

\newcommand\blfootnote[1]{%
  \begingroup
  \renewcommand\thefootnote{}\footnote{#1}%
  \addtocounter{footnote}{-1}%
  \endgroup
}

\usepackage{pgfplots}
\pgfplotsset{compat=1.17}
\usetikzlibrary{arrows.meta}
\usetikzlibrary{backgrounds}
\usepgfplotslibrary{patchplots}
\usepgfplotslibrary{fillbetween}
\pgfplotsset{%
    layers/standard/.define layer set={%
        background,axis background,axis grid,axis ticks,axis lines,axis tick labels,pre main,main,axis descriptions,axis foreground%
    }{
        grid style={/pgfplots/on layer=axis grid},%
        tick style={/pgfplots/on layer=axis ticks},%
        axis line style={/pgfplots/on layer=axis lines},%
        label style={/pgfplots/on layer=axis descriptions},%
        legend style={/pgfplots/on layer=axis descriptions},%
        title style={/pgfplots/on layer=axis descriptions},%
        colorbar style={/pgfplots/on layer=axis descriptions},%
        ticklabel style={/pgfplots/on layer=axis tick labels},%
        axis background@ style={/pgfplots/on layer=axis background},%
        3d box foreground style={/pgfplots/on layer=axis foreground},%
    },
}

\usepackage{algorithmicx}
\usepackage{algorithm}
\usepackage{algpseudocode}

\algdef{SE}{For}{EndFor}[1]{\textbf{for} \(\mbox{#1}\)}{\textbf{end}}%
\algdef{SE}[FUNCTION]{Function}{EndFunction}%
   [2]{\algorithmicfunction\ \textproc{#1}\ifthenelse{\equal{#2}{}}{}{(#2)}}%
   {\algorithmicend}%
\algtext*{EndFunction} 

\algrenewcommand\alglinenumber[1]{\color{gray}\tiny #1}

\definecolor{commentgray}{rgb}{0.7, 0.7, 0.7}
\newcommand{\GrayComment}[1]{{\hfill{\color{commentgray}$\triangleright$ #1}}}

\usepackage{listings}
\usepackage{xcolor}

\lstdefinelanguage{Julia}{
    keywords=[3]{abs,abs2,abspath,accept,accumulate,accumulate!,acos,acos_fast,acosd,acosh,acosh_fast,acot,acotd,acoth,acsc,acscd,acsch,adjoint,adjoint!,all,all!,allunique,angle,angle_fast,any,any!,append!,apropos,ascii,asec,asecd,asech,asin,asin_fast,asind,asinh,asinh_fast,assert,asyncmap,asyncmap!,atan,atan2,atan2_fast,atan_fast,atand,atanh,atanh_fast,atexit,atreplinit,axes,backtrace,base,basename,beta,big,bin,bind,binomial,bitbroadcast,bitrand,bits,bitstring,bkfact,bkfact!,blkdiag,broadcast,broadcast!,broadcast_getindex,broadcast_setindex!,bswap,bytes2hex,cat,catch_backtrace,catch_stacktrace,cbrt,cbrt_fast,cd,ceil,cfunction,cglobal,charwidth,checkbounds,checkindex,chmod,chol,cholfact,cholfact!,chomp,chop,chown,chr2ind,circcopy!,circshift,circshift!,cis,cis_fast,clamp,clamp!,cld,clipboard,close,cmp,coalesce,code_llvm,code_lowered,code_native,code_typed,code_warntype,codeunit,codeunits,collect,colon,complex,cond,condskeel,conj,conj!,connect,consume,contains,convert,copy,copy!,copysign,copyto!,cor,cos,cos_fast,cosc,cosd,cosh,cosh_fast,cospi,cot,cotd,coth,count,count_ones,count_zeros,countlines,countnz,cov,cp,cross,csc,cscd,csch,ctime,ctranspose,ctranspose!,cummax,cummin,cumprod,cumprod!,cumsum,cumsum!,current_module,current_task,dec,deepcopy,deg2rad,delete!,deleteat!,den,denominator,deserialize,det,detach,diag,diagind,diagm,diff,digits,digits!,dirname,disable_sigint,display,displayable,displaysize,div,divrem,done,dot,download,dropzeros,dropzeros!,dump,eachcol,eachindex,eachline,eachmatch,edit,eig,eigfact,eigfact!,eigmax,eigmin,eigvals,eigvals!,eigvecs,eltype,empty,empty!,endof,endswith,enumerate,eof,eps,equalto,error,esc,escape_string,evalfile,exit,exp,exp10,exp10_fast,exp2,exp2_fast,exp_fast,expanduser,expm,expm!,expm1,expm1_fast,exponent,extrema,eye,factorial,factorize,falses,fd,fdio,fetch,fieldcount,fieldname,fieldnames,fieldoffset,filemode,filesize,fill,fill!,filter,filter!,finalize,finalizer,find,findfirst,findin,findlast,findmax,findmax!,findmin,findmin!,findn,findnext,findnz,findprev,first,fld,fld1,fldmod,fldmod1,flipbits!,flipdim,flipsign,float,floor,flush,fma,foldl,foldr,foreach,frexp,full,fullname,functionloc,gamma,gc,gc_enable,gcd,gcdx,gensym,get,get!,get_zero_subnormals,getaddrinfo,getalladdrinfo,gethostname,getindex,getipaddr,getkey,getnameinfo,getpeername,getpid,getsockname,givens,gperm,gradient,hash,haskey,hcat,hessfact,hessfact!,hex,hex2bytes,hex2bytes!,hex2num,homedir,htol,hton,hvcat,hypot,hypot_fast,identity,ifelse,ignorestatus,im,imag,in,include_dependency,include_string,ind2chr,ind2sub,indexin,indices,indmax,indmin,info,insert!,instances,intersect,intersect!,inv,invmod,invperm,invpermute!,ipermute!,ipermutedims,is,is_apple,is_bsd,is_linux,is_unix,is_windows,isabspath,isapprox,isascii,isassigned,isbits,isblockdev,ischardev,isconcrete,isconst,isdiag,isdir,isdirpath,isempty,isequal,iseven,isfifo,isfile,isfinite,ishermitian,isimag,isimmutable,isinf,isinteger,isinteractive,isleaftype,isless,isletter,islink,islocked,ismarked,ismatch,ismissing,ismount,isnan,isodd,isone,isopen,ispath,isperm,isposdef,isposdef!,ispow2,isqrt,isreadable,isreadonly,isready,isreal,issetgid,issetuid,issocket,issorted,issparse,issticky,issubnormal,issubset,issubtype,issymmetric,istaskdone,istaskstarted,istextmime,istril,istriu,isvalid,iswritable,iszero,join,joinpath,keys,keytype,kill,kron,last,lbeta,lcm,ldexp,ldltfact,ldltfact!,leading_ones,leading_zeros,length,less,lexcmp,lexless,lfact,lgamma,lgamma_fast,linearindices,linreg,linspace,listen,listenany,lock,log,log10,log10_fast,log1p,log1p_fast,log2,log2_fast,log_fast,logabsdet,logdet,logging,logm,logspace,lpad,lq,lqfact,lqfact!,lstat,lstrip,ltoh,lu,lufact,lufact!,lyap,macroexpand,map,map!,mapfoldl,mapfoldr,mapreduce,mapreducedim,mapslices,mark,match,matchall,max,max_fast,maxabs,maximum,maximum!,maxintfloat,mean,mean!,median,median!,merge,merge!,method_exists,methods,methodswith,middle,midpoints,mimewritable,min,min_fast,minabs,minimum,minimum!,minmax,minmax_fast,missing,mkdir,mkpath,mktemp,mktempdir,mod,mod1,mod2pi,modf,module_name,module_parent,mtime,muladd,mv,names,nb_available,ncodeunits,ndigits,ndims,next,nextfloat,nextind,nextpow,nextpow2,nextprod,nnz,nonzeros,norm,normalize,normalize!,normpath,notify,ntoh,ntuple,nullspace,num,num2hex,numerator,nzrange,object_id,occursin,oct,oftype,one,ones,oneunit,open,operm,ordschur,ordschur!,pairs,parent,parentindexes,parentindices,parse,partialsort,partialsort!,partialsortperm,partialsortperm!,peakflops,permute,permute!,permutedims,permutedims!,pi,pinv,pipeline,pointer,pointer_from_objref,pop!,popdisplay,popfirst!,position,pow_fast,powermod,precision,precompile,prepend!,prevfloat,prevind,prevpow,prevpow2,print,print_shortest,print_with_color,println,process_exited,process_running,prod,prod!,produce,promote,promote_rule,promote_shape,promote_type,push!,pushdisplay,pushfirst!,put!,pwd,qr,qrfact,qrfact!,quantile,quantile!,quit,rad2deg,rand,rand!,randcycle,randcycle!,randexp,randexp!,randjump,randn,randn!,randperm,randperm!,randstring,randsubseq,randsubseq!,range,rank,rationalize,read,read!,readandwrite,readavailable,readbytes!,readchomp,readdir,readline,readlines,readlink,readstring,readuntil,real,realmax,realmin,realpath,recv,recvfrom,redirect_stderr,redirect_stdin,redirect_stdout,redisplay,reduce,reducedim,reenable_sigint,reim,reinterpret,reload,relpath,rem,rem2pi,repeat,replace,replace!,repmat,repr,reprmime,reset,reshape,resize!,rethrow,retry,reverse,reverse!,reverseind,rm,rol,rol!,ror,ror!,rot180,rotl90,rotr90,round,rounding,rowvals,rpad,rsearch,rsearchindex,rsplit,rstrip,run,scale!,schedule,schur,schurfact,schurfact!,search,searchindex,searchsorted,searchsortedfirst,searchsortedlast,sec,secd,sech,seek,seekend,seekstart,select,select!,selectperm,selectperm!,send,serialize,set_zero_subnormals,setdiff,setdiff!,setenv,setindex!,setprecision,setrounding,shift!,show,showall,showcompact,showerror,shuffle,shuffle!,sign,signbit,signed,signif,significand,similar,sin,sin_fast,sinc,sincos,sind,sinh,sinh_fast,sinpi,size,sizehint!,sizeof,skip,skipchars,skipmissing,sleep,slicedim,sort,sort!,sortcols,sortperm,sortperm!,sortrows,sparse,sparsevec,spawn,spdiagm,speye,splice!,split,splitdir,splitdrive,splitext,spones,sprand,sprandn,sprint,spzeros,sqrt,sqrt_fast,sqrtm,squeeze,srand,stacktrace,start,startswith,stat,std,stdm,step,stride,strides,string,stringmime,strip,strwidth,sub2ind,subtypes,success,sum,sum!,sumabs,sumabs2,summary,supertype,svd,svdfact,svdfact!,svdvals,svdvals!,sylvester,symdiff,symdiff!,symlink,systemerror,take!,takebuf_array,takebuf_string,tan,tan_fast,tand,tanh,tanh_fast,task_local_storage,tempdir,tempname,thisind,tic,time,time_ns,timedwait,to_indices,toc,toq,touch,trace,trailing_ones,trailing_zeros,transcode,transpose,transpose!,tril,tril!,triu,triu!,trues,trunc,truncate,trylock,tryparse,typeintersect,typejoin,typemax,typemin,unescape_string,union,union!,unique,unique!,unlock,unmark,unsafe_copy!,unsafe_copyto!,unsafe_load,unsafe_pointer_to_objref,unsafe_read,unsafe_store!,unsafe_string,unsafe_trunc,unsafe_wrap,unsafe_write,unshift!,unsigned,uperm,valtype,values,var,varinfo,varm,vcat,vec,vecdot,vecnorm,versioninfo,view,wait,walkdir,warn,which,whos,widemul,widen,withenv,workspace,write,xor,yield,yieldto,zero,zeros,zip,applicable,eval,fieldtype,getfield,invoke,isa,isdefined,nfields,nothing,setfield!,throw,tuple,typeassert,typeof,uninitialized},%
    keywords=[3]{asum,axpby!,axpy!,blascopy!,dot,dotc,dotu,gbmv,gbmv!,gemm,gemm!,gemv,gemv!,ger!,hemm,hemm!,hemv,hemv!,her!,her2k,her2k!,herk,herk!,iamax,nrm2,sbmv,sbmv!,scal,scal!,symm,symm!,symv,symv!,syr!,syr2k,syr2k!,syrk,syrk!,trmm,trmm!,trmv,trmv!,trsm,trsm!,trsv,trsv!),abs,abs2,abspath,accept,accumulate,accumulate!,acos,acos_fast,acosd,acosh,acosh_fast,acot,acotd,acoth,acsc,acscd,acsch,adjoint,adjoint!,all,all!,allunique,angle,angle_fast,any,any!,append!,apropos,argmax,argmin,ascii,asec,asecd,asech,asin,asin_fast,asind,asinh,asinh_fast,assert,asyncmap,asyncmap!,atan,atan2,atan2_fast,atan_fast,atand,atanh,atanh_fast,atexit,atreplinit,axes,backtrace,base,basename,beta,bfft,bfft!,big,bin,bind,binomial,bitbroadcast,bitrand,bits,bitstring,bkfact,bkfact!,blkdiag,brfft,broadcast,broadcast!,broadcast_getindex,broadcast_setindex!,bswap,bytes2hex,cat,catch_backtrace,catch_stacktrace,cbrt,cbrt_fast,cd,ceil,cfunction,cglobal,charwidth,checkbounds,checkindex,chmod,chol,cholfact,cholfact!,chomp,chop,chown,chr2ind,circcopy!,circshift,circshift!,cis,cis_fast,clamp,clamp!,cld,clipboard,close,cmp,coalesce,code_llvm,code_lowered,code_native,code_typed,code_warntype,codeunit,codeunits,collect,colon,complex,cond,condskeel,conj,conj!,connect,consume,contains,conv,conv2,convert,copy,copy!,copysign,copyto!,cor,cos,cos_fast,cosc,cosd,cosh,cosh_fast,cospi,cot,cotd,coth,count,count_ones,count_zeros,countlines,countnz,cov,cp,cross,csc,cscd,csch,ctime,ctranspose,ctranspose!,cummax,cummin,cumprod,cumprod!,cumsum,cumsum!,current_module,current_task,dct,dct!,dec,deconv,deepcopy,deg2rad,delete!,deleteat!,den,denominator,deserialize,det,detach,diag,diagind,diagm,diff,digits,digits!,dirname,disable_sigint,display,displayable,displaysize,div,divrem,done,dot,download,dropzeros,dropzeros!,dump,eachcol,eachindex,eachline,eachmatch,edit,eig,eigfact,eigfact!,eigmax,eigmin,eigvals,eigvals!,eigvecs,eltype,empty,empty!,endof,endswith,enumerate,eof,eps,equalto,error,esc,escape_string,evalfile,exit,exp,exp10,exp10_fast,exp2,exp2_fast,exp_fast,expand,expanduser,expm,expm!,expm1,expm1_fast,exponent,extrema,eye,factorial,factorize,falses,fd,fdio,fetch,fft,fft!,fftshift,fieldcount,fieldname,fieldnames,fieldoffset,filemode,filesize,fill,fill!,filt,filt!,filter,filter!,finalize,finalizer,find,findfirst,findin,findlast,findmax,findmax!,findmin,findmin!,findn,findnext,findnz,findprev,first,fld,fld1,fldmod,fldmod1,flipbits!,flipdim,flipsign,float,floor,flush,fma,foldl,foldr,foreach,frexp,full,fullname,functionloc,gamma,gc,gc_enable,gcd,gcdx,gensym,get,get!,get_zero_subnormals,getaddrinfo,getalladdrinfo,gethostname,getindex,getipaddr,getkey,getnameinfo,getpeername,getpid,getsockname,givens,gperm,gradient,hash,haskey,hcat,hessfact,hessfact!,hex,hex2bytes,hex2bytes!,hex2num,homedir,htol,hton,hvcat,hypot,hypot_fast,idct,idct!,identity,ifelse,ifft,ifft!,ifftshift,ignorestatus,im,imag,in,include_dependency,include_string,ind2chr,ind2sub,indexin,indices,indmax,indmin,info,insert!,instances,intersect,intersect!,inv,invmod,invperm,invpermute!,ipermute!,ipermutedims,irfft,is,is_apple,is_bsd,is_linux,is_unix,is_windows,isabspath,isapprox,isascii,isassigned,isbits,isblockdev,ischardev,isconcrete,isconst,isdiag,isdir,isdirpath,isempty,isequal,iseven,isfifo,isfile,isfinite,ishermitian,isimag,isimmutable,isinf,isinteger,isinteractive,isleaftype,isless,isletter,islink,islocked,ismarked,ismatch,ismissing,ismount,isnan,isodd,isone,isopen,ispath,isperm,isposdef,isposdef!,ispow2,isqrt,isreadable,isreadonly,isready,isreal,issetgid,issetuid,issocket,issorted,issparse,issticky,issubnormal,issubset,issubtype,issymmetric,istaskdone,istaskstarted,istextmime,istril,istriu,isvalid,iswritable,iszero,join,joinpath,keys,keytype,kill,kron,last,lbeta,lcm,ldexp,ldltfact,ldltfact!,leading_ones,leading_zeros,length,less,lexcmp,lexless,lfact,lgamma,lgamma_fast,linearindices,linreg,linspace,listen,listenany,lock,log,log10,log10_fast,log1p,log1p_fast,log2,log2_fast,log_fast,logabsdet,logdet,logging,logm,logspace,lpad,lq,lqfact,lqfact!,lstat,lstrip,ltoh,lu,lufact,lufact!,lyap,macroexpand,map,map!,mapfoldl,mapfoldr,mapreduce,mapreducedim,mapslices,mark,match,matchall,max,max_fast,maxabs,maximum,maximum!,maxintfloat,mean,mean!,median,median!,merge,merge!,method_exists,methods,methodswith,middle,midpoints,mimewritable,min,min_fast,minabs,minimum,minimum!,minmax,minmax_fast,missing,mkdir,mkpath,mktemp,mktempdir,mod,mod1,mod2pi,modf,module_name,module_parent,mtime,muladd,mv,names,nb_available,ncodeunits,ndigits,ndims,next,nextfloat,nextind,nextpow,nextpow2,nextprod,nnz,nonzeros,norm,normalize,normalize!,normpath,notify,ntoh,ntuple,nullspace,num,num2hex,numerator,nzrange,object_id,occursin,oct,oftype,one,ones,oneunit,open,operm,ordschur,ordschur!,pairs,parent,parentindexes,parentindices,parse,partialsort,partialsort!,partialsortperm,partialsortperm!,peakflops,permute,permute!,permutedims,permutedims!,pi,pinv,pipeline,plan_bfft,plan_bfft!,plan_brfft,plan_dct,plan_dct!,plan_fft,plan_fft!,plan_idct,plan_idct!,plan_ifft,plan_ifft!,plan_irfft,plan_rfft,pointer,pointer_from_objref,pop!,popdisplay,popfirst!,position,pow_fast,powermod,precision,precompile,prepend!,prevfloat,prevind,prevpow,prevpow2,print,print_shortest,print_with_color,println,process_exited,process_running,prod,prod!,produce,promote,promote_rule,promote_shape,promote_type,push!,pushdisplay,pushfirst!,put!,pwd,qr,qrfact,qrfact!,quantile,quantile!,quit,rad2deg,rand,rand!,randcycle,randcycle!,randexp,randexp!,randjump,randn,randn!,randperm,randperm!,randstring,randsubseq,randsubseq!,range,rank,rationalize,read,read!,readandwrite,readavailable,readbytes!,readchomp,readdir,readline,readlines,readlink,readstring,readuntil,real,realmax,realmin,realpath,recv,recvfrom,redirect_stderr,redirect_stdin,redirect_stdout,redisplay,reduce,reducedim,reenable_sigint,reim,reinterpret,reload,relpath,rem,rem2pi,repeat,replace,replace!,repmat,repr,reprmime,reset,reshape,resize!,rethrow,retry,reverse,reverse!,reverseind,rfft,rm,rol,rol!,ror,ror!,rot180,rotl90,rotr90,round,rounding,rowvals,rpad,rsearch,rsearchindex,rsplit,rstrip,run,scale!,schedule,schur,schurfact,schurfact!,search,searchindex,searchsorted,searchsortedfirst,searchsortedlast,sec,secd,sech,seek,seekend,seekstart,select,select!,selectperm,selectperm!,send,serialize,set_zero_subnormals,setdiff,setdiff!,setenv,setindex!,setprecision,setrounding,shift!,show,showall,showcompact,showerror,shuffle,shuffle!,sign,signbit,signed,signif,significand,similar,sin,sin_fast,sinc,sincos,sind,sinh,sinh_fast,sinpi,size,sizehint!,sizeof,skip,skipchars,skipmissing,sleep,slicedim,sort,sort!,sortcols,sortperm,sortperm!,sortrows,sparse,sparsevec,spawn,spdiagm,speye,splice!,split,splitdir,splitdrive,splitext,spones,sprand,sprandn,sprint,spzeros,sqrt,sqrt_fast,sqrtm,squeeze,srand,stacktrace,start,startswith,stat,std,stdm,step,stride,strides,string,stringmime,strip,strwidth,sub2ind,subtypes,success,sum,sum!,sumabs,sumabs2,summary,super,supertype,svd,svdfact,svdfact!,svdvals,svdvals!,sylvester,symdiff,symdiff!,symlink,systemerror,take!,takebuf_array,takebuf_string,tan,tan_fast,tand,tanh,tanh_fast,task_local_storage,tempdir,tempname,thisind,tic,time,time_ns,timedwait,to_indices,toc,toq,touch,trace,trailing_ones,trailing_zeros,transcode,transpose,transpose!,tril,tril!,triu,triu!,trues,trunc,truncate,trylock,tryparse,typeintersect,typejoin,typemax,typemin,unescape_string,union,union!,unique,unique!,unlock,unmark,unsafe_copy!,unsafe_copyto!,unsafe_load,unsafe_pointer_to_objref,unsafe_read,unsafe_store!,unsafe_string,unsafe_trunc,unsafe_wrap,unsafe_write,unshift!,unsigned,uperm,valtype,values,var,varinfo,varm,vcat,vec,vecdot,vecnorm,versioninfo,view,wait,walkdir,warn,which,whos,widemul,widen,withenv,workspace,write,xcorr,xor,yield,yieldto,zero,zeros,zip,broadcast_getindex,broadcast_indices,broadcast_setindex!,broadcast_similar,dotview,apropos,doc,countfrom,cycle,drop,enumerate,flatten,partition,product,repeated,rest,take,zip,get_creds!,with,calloc,errno,flush_cstdio,free,gethostname,getpid,malloc,realloc,strerror,strftime,strptime,systemsleep,time,transcode,dlclose,dlext,dllist,dlopen,dlopen_e,dlpath,dlsym,dlsym_e,find_library,adjoint,adjoint!,axpby!,axpy!,bkfact,bkfact!,chol,cholfact,cholfact!,cond,condskeel,copy_transpose!,copyto!,cross,det,diag,diagind,diagm,diff,dot,eig,eigfact,eigfact!,eigmax,eigmin,eigvals,eigvals!,eigvecs,factorize,getq,givens,gradient,hessfact,hessfact!,isdiag,ishermitian,isposdef,isposdef!,issuccess,issymmetric,istril,istriu,kron,ldltfact,ldltfact!,linreg,logabsdet,logdet,lq,lqfact,lqfact!,lu,lufact,lufact!,lyap,norm,normalize,normalize!,nullspace,ordschur,ordschur!,peakflops,pinv,qr,qrfact,qrfact!,rank,scale!,schur,schurfact,schurfact!,svd,svdfact,svdfact!,svdvals,svdvals!,sylvester,trace,transpose,transpose!,transpose_type,tril,tril!,triu,triu!,vecdot,vecnorm,html,latex,license,readme,isexpr,quot,show_sexpr,add,available,build,checkout,clone,dir,free,init,installed,pin,resolve,rm,setprotocol!,status,test,update,deserialize,serialize,blkdiag,droptol!,dropzeros,dropzeros!,issparse,nnz,nonzeros,nzrange,permute,rowvals,sparse,sparsevec,spdiagm,spones,sprand,sprandn,spzeros,catch_stacktrace,stacktrace,cpu_info,cpu_summary,free_memory,isapple,isbsd,islinux,isunix,iswindows,loadavg,total_memory,uptime,atomic_add!,atomic_and!,atomic_cas!,atomic_fence,atomic_max!,atomic_min!,atomic_nand!,atomic_or!,atomic_sub!,atomic_xchg!,atomic_xor!,nthreads,threadid,applicable,eval,fieldtype,getfield,invoke,isa,isdefined,nfields,nothing,setfield!,throw,tuple,typeassert,typeof,uninitialized},%
    keywords=[2]{AbstractArray,AbstractChannel,AbstractDict,AbstractDisplay,AbstractFloat,AbstractIrrational,AbstractMatrix,AbstractRNG,AbstractRange,AbstractSerializer,AbstractSet,AbstractSparseArray,AbstractSparseMatrix,AbstractSparseVector,AbstractString,AbstractUnitRange,AbstractVecOrMat,AbstractVector,Adjoint,Any,ArgumentError,Array,AssertionError,Bidiagonal,BigFloat,BigInt,BitArray,BitMatrix,BitSet,BitVector,Bool,BoundsError,BufferStream,CapturedException,CartesianIndex,CartesianIndices,Cchar,Cdouble,Cfloat,Channel,Char,Cint,Cintmax_t,Clong,Clonglong,Cmd,CodeInfo,Colon,Complex,ComplexF16,ComplexF32,ComplexF64,CompositeException,Condition,ConjArray,ConjMatrix,ConjVector,Cptrdiff_t,Cshort,Csize_t,Cssize_t,Cstring,Cuchar,Cuint,Cuintmax_t,Culong,Culonglong,Cushort,Cvoid,Cwchar_t,Cwstring,DataType,DenseArray,DenseMatrix,DenseVecOrMat,DenseVector,Diagonal,Dict,DimensionMismatch,Dims,DivideError,DomainError,EOFError,EachLine,Enum,Enumerate,ErrorException,Exception,ExponentialBackOff,Expr,Factorization,Float16,Float32,Float64,Function,GlobalRef,GotoNode,HTML,Hermitian,IO,IOBuffer,IOContext,IOStream,IPAddr,IPv4,IPv6,IndexCartesian,IndexLinear,IndexStyle,InexactError,InitError,Int,Int128,Int16,Int32,Int64,Int8,Integer,InterruptException,InvalidStateException,Irrational,KeyError,LabelNode,LinSpace,LineNumberNode,LinearIndices,LoadError,LowerTriangular,MIME,Matrix,MersenneTwister,Method,MethodError,MethodTable,Missing,MissingException,Module,NTuple,NamedTuple,NewvarNode,Nothing,Number,ObjectIdDict,OrdinalRange,OutOfMemoryError,OverflowError,Pair,PartialQuickSort,PermutedDimsArray,Pipe,Ptr,QuoteNode,RandomDevice,Rational,RawFD,ReadOnlyMemoryError,Real,ReentrantLock,Ref,Regex,RegexMatch,RoundingMode,RowVector,SSAValue,SegmentationFault,SerializationState,Set,Signed,SimpleVector,Slot,SlotNumber,Some,SparseMatrixCSC,SparseVector,StackFrame,StackOverflowError,StackTrace,StepRange,StepRangeLen,StridedArray,StridedMatrix,StridedVecOrMat,StridedVector,String,StringIndexError,SubArray,SubString,SymTridiagonal,Symbol,Symmetric,SystemError,TCPSocket,Task,Text,TextDisplay,Timer,Transpose,Tridiagonal,Tuple,Type,TypeError,TypeMapEntry,TypeMapLevel,TypeName,TypeVar,TypedSlot,UDPSocket,UInt,UInt128,UInt16,UInt32,UInt64,UInt8,UndefRefError,UndefVarError,UniformScaling,Uninitialized,Union,UnionAll,UnitRange,Unsigned,UpperTriangular,Val,Vararg,VecElement,VecOrMat,Vector,VersionNumber,WeakKeyDict,WeakRef,BLAS,Base,Broadcast,DFT,Docs,Iterators,LAPACK,LibGit2,Libc,Libdl,LinAlg,Markdown,Meta,Operators,Pkg,Serializer,SparseArrays,StackTraces,Sys,Threads,Core,Main},%
    keywords=[1]{true,false,nothing,missing,im,uninitialized,NaN,NaN16,NaN32,NaN64,Inf,Inf16,Inf32,Inf64,ARGS,C_NULL,ENDIAN_BOM,ENV,LOAD_PATH,PROGRAM_FILE,STDERR,STDIN,STDOUT,VERSION},
    keywords=[1]{mutable,immutable,struct,begin,end,function,macro,quote,let,local,global,const,abstract,module,baremodule,using,import,export,in,if,else,elseif,for,while,do,try,type,catch,finally,return,break,continue},%
    sensitive=true,
    morecomment=[l]{\#},
    morecomment=[n]{\#=}{=\#},
    morestring=[s]{"}{"},
    morestring=[m]{'}{'},
    alsoletter=!?
}

\lstdefinestyle{julia}{
    backgroundcolor  = \color[HTML]{F2F2F2},
    basicstyle       = \ttfamily\footnotesize\color[HTML]{19177C},
    numberstyle      = \ttfamily\scriptsize\color[HTML]{7F7F7F},
    keywordstyle     = [1]{\bfseries\color[HTML]{1BA1EA}},
    keywordstyle     = [2]{\color[HTML]{0F6FA3}},
    keywordstyle     = [3]{\color[HTML]{0000FF}},
    stringstyle      = \ttfamily\color[HTML]{F5615C},
    commentstyle     = \color[HTML]{AAAAAA},
    rulecolor        = \color[HTML]{000000},
    frame=lines,
    xleftmargin=15pt,
    framexleftmargin=15pt,
    framextopmargin=4pt,
    framexbottommargin=4pt,
    tabsize=4,
    captionpos=b,
    breaklines=true,
    breakatwhitespace=false,
    showstringspaces=false,
    showspaces=false,
    showtabs=false,
    columns=fullflexible,
    keepspaces=true,
    numbers=none
}

\lstdefinelanguage{JuliaLocal}{
    language = Julia, 
    morekeywords = [3]{reset, initialize, generate_input, evaluate, bayesian_safety_validation, falsification, most_likely_failure, p_estimate}, 
    morekeywords = [2]{BayesianSafetyValidation, SystemParameters, Input, RunwayDetectionSystemParameters, OperationalParameters, TruncatedNormal, Normal}, 
}

\title{Bayesian Safety Validation for Failure Probability\\Estimation of Black-Box Systems}

\author{%
Robert J. Moss\footnote{Corresponding Author, PhD Student, Department of Computer Science, Stanford, CA, AIAA Student Member, \texttt{mossr@cs.stanford.edu}}
and
Mykel J. Kochenderfer\footnote{Associate Professor, Department of Aeronautics and Astronautics, Stanford, CA, AIAA Associate Fellow, \texttt{mykel@stanford.edu}}}
\affil{Stanford University, Stanford, CA, 94305}

\author{%
Maxime Gariel\footnote{Chief Technology Officer, San Francisco, CA, \texttt{maxime@xwing.com}}
and
Arthur Dubois\footnote{Director of Engineering, San Francisco, CA, \texttt{arthur.dubois@xwing.com}}}
\affil{Xwing, San Francisco, CA, 94102}

\begin{document}

\blfootnote{Presented as Paper 2023-3596 at the 2023 AIAA AVIATION Forum, San Diego, CA, 12--16 June 2023.}

\maketitle

\begin{singlespacing}

\begin{abstract}
Estimating the probability of failure is an important step in the certification of safety-critical systems.
Efficient estimation methods are often needed due to the challenges posed by high-dimensional input spaces, risky test scenarios, and computationally expensive simulators.
This work frames the problem of black-box safety validation as a Bayesian optimization problem and introduces a method that iteratively fits a probabilistic surrogate model to efficiently predict failures.
The algorithm is designed to search for failures, compute the most-likely failure, and estimate the failure probability over an operating domain using importance sampling.
We introduce three acquisition functions that aim to reduce uncertainty by covering the design space, optimize the analytically derived failure boundaries, and sample the predicted failure regions.
Results show this Bayesian safety validation approach provides a more accurate estimate of failure probability with orders of magnitude fewer samples and performs well across various safety validation metrics.
We demonstrate this approach on three test problems, a stochastic decision making system, and a neural network-based runway detection system.
This work is open sourced (\url{https://github.com/sisl/BayesianSafetyValidation.jl}) and currently being used to supplement the FAA certification process of the machine learning components for an autonomous cargo aircraft.
\end{abstract}

\section*{Nomenclature}

{\renewcommand\arraystretch{1.0}
\noindent\begin{longtable*}{@{}l @{\quad=\quad} l@{}}
$f$  & failure indicator for the black-box system under test \\
$\hat{f}$ & predicted mean of probabilistic surrogate model \\
$\hat{\sigma}$ & predicted standard deviation of probabilistic surrogate model \\
$\hat{g}$ & surrogate model binary failure classification \\
$p$ & operational likelihood model (target/nominal distribution) \\
$q$ & importance sampling proposal distribution \\
$\vec{x}$& input vector from the design space \\
\end{longtable*}}

\section{Introduction}

\lettrine{C}{ertifying} safety-critical autonomous systems is an important step for their safe deployment in aviation. 
Examples of safety-critical systems include those for detect and avoid \cite{do365,owen2019acasxu}, collision avoidance \cite{kochenderfer2012next}, runway detection \cite{khaled2015runway}, and auto-land \cite{balduzzi2021neural}.
One way to provide a quantitative measure of safety is to estimate the probability of system failure.
The process of estimating the probability of failure can highlight areas of weakness in the system (by uncovering failures) and can show how well the system performs in their operating environments.
The rarity of failures makes it challenging to accurately estimate failure probability especially when using computationally expensive simulators \cite{de2005tutorial}.
Therefore, it is important to efficiently sample the design space when searching for failures (using a minimum set of inputs) and to maximize a measure of confidence in the resulting failure probability estimate.
A standard approach to estimating this rare-event probability involves Monte Carlo (MC) sampling to generate a set of system inputs from a likelihood model of the operating environment.
Estimating this rare-event probability through Monte Carlo sampling can be computationally expensive and usually requires many samples to minimize the variance of the estimate \cite{de2005tutorial}.
A variance-reduction technique to more efficiently estimate the failure probability uses \textit{importance sampling} \cite{robert1999monte,owen2013monte} to draw samples from a different distribution, called the proposal, and then re-weight the expectation based on the likelihood ratio between the operational model and the proposal.
Importance sampling is especially useful in the safety-critical case due to the unbiased failure probability estimate \cite{owen2013monte}.

Bayesian optimization algorithms such as the cross-entropy method (CEM) \cite{rubinstein2004cross,moss2020crossentropy} have been adapted to the problem of rare-event estimation through a multi-level procedure \cite{de2005tutorial, miller2021rare}, but rely on a real-valued system output with a defined failure threshold to adaptively narrow the search.
\citet{arief2021deep} proposed a deep importance sampling approach for rare-event estimation of black-box systems (Deep-PrAE) but rely on similar real-valued systems.
In our problem, the system under test outputs a binary value indicating failure and thus cannot effectively use these methods.
Population-based methods, like population Monte Carlo (PMC) \cite{cappe2004population} and optimized population Monte Carlo (O-PMC) \cite{elvira2022optimized}, 
work well for both real-valued and binary-valued systems and use adaptive importance sampling \cite{bugallo2017adaptive} to iteratively estimate the optimal proposal distribution.
The PMC algorithms use self-normalized importance sampling (SNIS) to estimate the probability in question \cite{owen2013monte}.
Population-based approaches often require a large number of system evaluations to adequately converge (see \citet{luengo2020survey} for a comprehensive survey of Monte Carlo estimation algorithms).
\citet{vazquez2009sequential} and \citet{wang2016gaussian} consider the problem of failure probability estimation when dealing with computationally expensive systems.
They fit a Gaussian process surrogate model to the underlying real-valued system (i.e., not the system output indication of failure) and then estimate the failure probability over this surrogate, similar to work from \citet{renganathan2022multifidelity} for the multifidelity case.
Those methods may not work on binary-valued systems or scale to complex systems such as image-based neural networks.
\citet{he2020framework} propose a framework for analyzing safety-critical deep neural networks using Bayesian statistics to iteratively fit a decision boundary from a predefined dictionary of shapes.
They use a boundary acquisition function that is based on expected improvement \cite{ranjan2008sequential}, requiring a definition of an $\epsilon$-threshold around the predicted boundaries at $0.5 \pm \epsilon$.
Our proposed approach constructs the probabilistic surrogate model so that a failure boundary can be analytically derived.

With the goal of sample efficiency, this work reformulates the safety validation problem \cite{corso2021survey} as a Bayesian optimization problem \cite{mockus1989global,optbook,garnett_bayesoptbook_2023} and introduces a set of acquisition functions each with their own safety validation objective.
Applying a Bayesian approach allows us to fit a probabilistic surrogate model to a minimal set of design points evaluated from the true system and then estimate failure probability using importance sampling on the inexpensive surrogate.
As a real-world case study, we use the proposed algorithm to estimate the probability of failure for a neural network-based runway detection system where the design space consists of the glide slope angle and the distance to runway.
The parametric design space is used to generate an input image of a runway in simulation, conditioned on the knowledge that the aircraft is in an approach, and the output is a binary value of failure (i.e., a misdetection).

The goals of this work are to: 
(1) estimate the probability of failure for a black-box safety-critical subsystem,
(2) focus on sample efficiency using the minimal number of data points,
(3) find realistic cases using a model of the environment the system will be operating in, weighting the failures based on their operational likelihood,
(4) characterize the entire set of failure regions to identify model weaknesses for further development, and
(5) ensure the entire design space is adequately covered.
The proposed \textit{Bayesian safety validation} (BSV) algorithm can be applied to general black-box systems to find failures, determine the most-likely failure, and estimate the overall failure probability.
An open-source Julia framework\footnote{\url{https://github.com/sisl/BayesianSafetyValidation.jl}} was developed to extend this work to other black-box systems and reproduce the results in this paper.

\section{Background}\label{sec:background}

To understand the methods developed in this work, we will provide the necessary background by first introducing the problem of safety validation and then will briefly discuss Gaussian processes and their use in Bayesian optimization.

\subsection{Safety Validation}\label{sec:safety}

Safety validation has three primary tasks \cite{corso2021survey} shown in \cref{fig:safety_validation}.
The first task, \textit{falsification}, is the process of finding any input that results in system failure.
The second task, \textit{most-likely failure analysis}, tries to find the failures with maximum likelihood.
And the third task, \textit{failure probability estimation}, estimates the probability that a failure will occur.
In focusing on failure probability estimation, we can achieve all three safety validation tasks. This is because when we estimate the probability of failure, we generate a distribution of failures.
Thus, we achieve falsification by finding failures in the process of constructing the distribution and can easily compute the most-likely failure by maximizing the input likelihood across the distribution.
Motivated to achieve all three safety validation tasks, this work develops an efficient approach to estimate probability of failure for black-box systems.
For a survey of existing black-box safety validation algorithms, including falsification and most-likely failure analysis, we refer to \citet{corso2021survey}.

\begin{figure}[t]
    \centering
    \resizebox{\textwidth}{!}{%
          \tikzset{
    >={Latex[width=1.5mm,length=1.5mm]},
    end/.style = {circle, minimum width=2mm, minimum height=2mm},
    startstyle/.style = {rectangle, semithick, fill=white, draw=black, minimum width=3mm, minimum height=3mm},
    successstyle/.style = {end,  fill=darkgreen},
    failurestyle/.style = {end, fill=darkred},
    sectionstyle/.style = {fill=none, draw=none, minimum height=2cm},
}


\begin{tikzpicture}

\node (falsification) [sectionstyle, label={above:\textsc{Falsification}}] {
\begin{tikzpicture}
    \node (start) [startstyle] {};

    \node (success) [successstyle, above right=3mm and 25mm of start, label={right:{\color{darkgreen}success}}] {};
    \node (failure) [failurestyle, below right=3mm and 25mm of start, label={right:{\color{darkred}failure}}] {};

    \draw [semithick,->] plot [smooth] coordinates {(start.east) ($(start.east)+(3mm,0mm)$)  ($(start.east)+(6mm,-3mm)$) ($(start.east)+(13mm,1mm)$) ($(start.east)+(20mm,-10mm)$) (failure.south west)};
\end{tikzpicture}
};

\draw [gray] ($(falsification.east) + (0.4cm,-2)$) -- ($(falsification.east) + (0.4cm,2)$);

\node (mostlikely) [sectionstyle, right=of falsification, label={[align=center]above:\textsc{Most-likely}\\\textsc{Failure Analysis}}] {
\begin{tikzpicture}
    \node (start) [startstyle] {};
    
    \node (success) [successstyle, above right=3mm and 25mm of start, label={right:{\color{darkgreen}success}}] {};   
    \node (failure) [failurestyle, below right=3mm and 25mm of start, label={right:{\color{darkred}failure}}] {};

    \draw [semithick,->] plot [smooth] coordinates {(start.east) ($(start.east)+(10mm,1mm)$) (failure.north west)};
\end{tikzpicture}
};

\draw [gray] ($(mostlikely.east) + (0.4cm,-2)$) -- ($(mostlikely.east) + (0.4cm,2)$);

\node (failureprob) [sectionstyle, right=of mostlikely, label={[align=center]above:\textsc{Failure Probability}\\\textsc{Estimation}}] {
\begin{tikzpicture}
    \node (start) [startstyle] {};
    
    \node (success) [successstyle, above right=3mm and 25mm of start, label={right:{\color{darkgreen}success}}] {};
    \node (failure) [failurestyle, below right=3mm and 25mm of start, label={right:{\color{darkred}failure}}] {};

    \draw [semithick,gray4,->] plot [smooth] coordinates {(start.east) ($(start.east)+(3mm,0mm)$)  ($(start.east)+(6mm,-3mm)$) ($(start.east)+(13mm,1mm)$) ($(start.east)+(20mm,-10mm)$) (failure.south west)};
    \draw [semithick,gray4,->] plot [smooth] coordinates {(start.east) ($(start.east)+(5mm,-2mm)$) ($(start.east)+(16mm,-12mm)$) (failure.south)};
    \draw [semithick,gray3,->] plot [smooth] coordinates {(start.east) ($(start.east)+(15mm,6mm)$) (failure.north)};
    \draw [semithick,gray2,->] plot [smooth] coordinates {(start.east) ($(start.east)+(5mm,-1mm)$) ($(start.east)+(15mm,-9mm)$) (failure.south west)};
    \draw [semithick,gray1,->] plot [smooth] coordinates {(start.east) ($(start.east)+(6mm,-1mm)$) ($(start.east)+(15mm,-6mm)$) (failure.west)};
    \draw [semithick,->] plot [smooth] coordinates {(start.east) ($(start.east)+(10mm,1mm)$) (failure.north west)};
\end{tikzpicture}
};

\end{tikzpicture}
    }
    \caption{The three tasks of safety validation.}
    \label{fig:safety_validation}
\end{figure}

In the case of \textit{black-box safety validation}, we treat the system $f$ as a ``black box'' and attempt to perform the three tasks described above.
The black-box assumption means that the only way to interact with the system is by passing inputs $\vec{x}$ and observing outputs $y = f(\vec{x})$.
This is in contrast to \textit{white-box validation} which requires information about the internals of the system to prove properties of safety \cite{Schumann2001,Fitting2012,Clarke2018}.
In choosing to perform black-box validation, we can apply the developed methods to more general systems, particularly to systems with neural network components.
Although recent work has focused on verifying deep neural networks \cite{liu2021algorithms,sidrane2022overt}, scaling to large networks remains a challenge.

\subsection{Bayesian Optimization and Probabilistic Surrogate Models}

The basic optimization problem \cite{optbook} is to maximize (or minimize) a real-valued function $f: \mathbb{R}^n \to \mathbb{R}$ subject to $\vec{x}$ lying in the design space $\mathcal{X} \subseteq \mathbb{R}^n$:
\begin{align}
    \operatorname*{maximize}_{\vec{x}} \quad& f(\vec{x})\\
    \textrm{subject to} \quad& \vec{x} \in \mathcal{X}\nonumber
\end{align}

Bayesian optimization is a black-box approach to globally optimize the objective $f$ without requiring any information about internals of the function, e.g., no requirement on gradient information \cite{frazier2018tutorial,garnett_bayesoptbook_2023}.
The main idea is to iteratively fit a \textit{probabilistic surrogate model}---such as a Gaussian process \cite{williams2006gaussian}---to evaluation points of the true objective function and then propose new design points to evaluate based on the information and uncertainty quantified in the surrogate.
Bayesian optimization is especially useful when $f$ is computationally expensive to evaluate and the surrogate $\hat{f}$ is fast to evaluate in comparison \cite{frazier2018tutorial}.
\Cref{fig:bayesian_optimization} illustrates a Bayesian optimization example where the next sampled design point $\vec{x}'$ (shown as a green triangle) maximizes the \textit{upper-confidence bound} (UCB) acquisition function \cite{optbook}:
\begin{equation}
    \vec{x}' = \argmax_{\vec{x} \in \mathcal{X}} \hat{f}(\vec{x}) + \lambda\hat{\sigma}(\vec{x})
\end{equation}
where $\hat{f}$ is the mean of the surrogate model, $\hat{\sigma}$ is the standard deviation, and $\lambda \ge 0$ controls the trade-off between exploration (based on the uncertainty) and exploitation (based on the mean). Using a probabilistic approach when fitting the surrogate model allows us to use uncertainty in the underlying objective when acquiring subsequent samples.

\subsubsection{Gaussian Processes}\label{sec:gp}

One method for constructing a probabilistic surrogate model is to use a \textit{Gaussian process} (GP) \cite{williams2006gaussian}. Given true observations from the objective function, a GP is defined as a distribution over possible underlying functions that describe the observations \cite{optbook} (illustrated in \cref{fig:bayesian_optimization} as purple dashed lines showing five functions sampled from the GP). 
Given the set of $n$ inputs $\vec{X}=\{\vec{x}_1, \ldots, \vec{x}_n\}$ and $n$ true observations $\vec{y} = [y_1, \ldots, y_n]$ where $y_i = f(\vec{x}_i)$, a Gaussian process is parameterized by a mean function $\vec{m}(\vec{X})$, generally set to the zero-mean function $\vec{m}(\vec{X})_i = m(\vec{x}_i)=0$ if no prior information is given, and a kernel function $\vec{K}(\vec{X},\vec{X})$ that captures the correlations between data points as covariances.
The output of the kernel function is an $n \times n$ matrix where the element $\vec{K}(\vec{X},\vec{X})_{i,j} = k(\vec{x}_i, \vec{x}_j)$. The kernel $k(\vec{x}_i, \vec{x}_j)$ may be selected based on spatial information about the relationship of neighboring data points in design space (e.g., if the relationship is smooth, then one can use a squared exponential kernel \cite{optbook}). In this work, we use the isotropic Mat\'ern $1/2$ kernel \cite{marc2002matern} with length scale $\ell=\exp(-1/10)$ and signal standard deviation $s_\sigma = \exp(-1/10)$:
\begin{equation}
    k(\vec{x}_i, \vec{x}_j) = s_{\sigma}^2 \exp\left(-|\vec{x}_i - \vec{x}_j| / \ell\right)
\end{equation}
The choice of kernel and its parameters can be separately optimized depending on the problem (see \citet{williams2006gaussian}), where the kernel was chosen for this work based on the characteristic that the Mat\'ern kernel can capture more variation in neighboring values \cite{williams2006gaussian}.
Using the mean function and kernel parameterization and conditioning on the true observations $\vec{y}$, the GP produces samples for new points $\vec{X}'$ of the function it is trying to estimate as 
\begin{gather}
    \hat{\vec{y}} \mid \vec{y} \sim \mathcal{N}\bigl(\vec{\mu}(\vec{X},\vec{X}',\vec{y}), \vec{\Sigma}(\vec{X},\vec{X}')\bigr) \\
    \vec{\mu}(\vec{X},\vec{X}',\vec{y}) = \vec{m}(\vec{X}') + \vec{K}(\vec{X}', \vec{X})\vec{K}(\vec{X},\vec{X})^{-1}(\vec{y} - \vec{m}(\vec{X}))\\
    \vec{\Sigma}(\vec{X},\vec{X}') = \vec{K}(\vec{X}', \vec{X}') - \vec{K}(\vec{X}', \vec{X})\vec{K}(\vec{X},\vec{X})^{-1}\vec{K}(\vec{X}, \vec{X}')
\end{gather}
where $\vec{\mu}$ and $\vec{\Sigma}$ are the mean and covariance functions.
Across the domain $\vec{X}'$, these estimate $\hat{\vec{y}}$ can now be used as surrogates for the true function $f(\vec{x}')$ for $\vec{x}' \in \vec{X}'$.

\begin{figure}[t!]
    \centering
    \resizebox{0.85\textwidth}{!}{%
        \input{figures/bayesian_optimization}
    }
    \caption{An example maximization problem using Gaussian process Bayesian optimization with UCB exploration.}
    \label{fig:bayesian_optimization}
\end{figure}

\paragraph{Predicting a probability with a Gaussian process.}
Because our system $f$ returns discrete values in $\{0,1\}$ and we want to predict a real-valued probability in $[0,1]$, we consider this a binary classification problem \cite{williams1998bayesian,nickisch2008approximations}.
We construct the GP to predict the logits $\hat{\vec{z}}$ (which we naturally define with zero mean to indicate no prior knowledge about failures) and then apply the logistic function (i.e., inverse logit or sigmoid) to get the predictions $\hat{\vec{y}}$:
\begin{gather}
    \hat{\vec{z}} \mid \operatorname{logit}(\vec{y}) \sim \mathcal{N}\bigl(\vec{\mu}(\vec{X},\vec{X}',\operatorname{logit}(\vec{y})), \vec{\Sigma}(\vec{X},\vec{X}')\bigr)\\
    \operatorname{logit}(y_i) = \log\left(\frac{\phi(y_i)}{1 - \phi(y_i)}\right)/s\\
    \hat{\vec{y}} = \phi^{-1}\Bigl(\operatorname{logit}^{-1}(\hat{\vec{z}})\Bigr) = \phi^{-1}\left(\frac{1}{1 + \exp(-s\hat{\vec{z}})}\right)\label{eq:sigmoid}
\end{gather}
where $\phi(y_i) = y_i(1 - \epsilon) + (1 - y_i)\epsilon$ and $\phi^{-1}(\hat{y}_i) = (\hat{y}_i - \epsilon) / (1 - 2\epsilon)$ to ensure well defined logits and $s$ controls the steepness of the sigmoid curve.
This construction can still be used even if $f$ already outputs values in $[0,1]$ instead of binary indicators; the GP will fit directly to the provided failure probability of each point.
When the output is binary, applying the logit transformations ensure that the prediction lies in $[0,1]$ and can be interpreted probabilistically.
Other approaches to predict a probability using a Gaussian process explore the case where $f$ is bounded and can be modeled as a Beta distribution \cite{jensen2013bounded}.
The logit approach allows us to analytically compute failure boundaries.

\section{Problem Formulation}\label{sec:problem}

We frame the black-box safety validation problem as a Bayesian optimization problem and use a Gaussian process surrogate model to predict failures.
Bayesian optimization is a natural approach to optimize some function $f: \mathbb{R}^n \to \mathbb{R}$, e.g., a black-box system.
But our problem uses a function $f: \mathbb{R}^n \to \mathbb{B}$, where $\mathbb{B}$ represents the Boolean domain (returning \texttt{true} for failures and \texttt{false} for non-failures, which can also be interpreted as $1$ and $0$, respectively).
Instead of maximizing or minimizing $f$, we frame the problem to find failure regions through exploration, refine failure boundaries, and refine likely failure regions through sampling the theoretically optimal failure distribution \cite{kahn1953methods,murphy2012machine,owen2013monte}.
We introduce a set of three acquisition functions that accomplish these objectives and call the acquisition procedure \textit{failure search and refinement} (FSAR), shown together in \cref{alg:fsar}.
Although we are primarily interested in the more restrictive case where $f$ outputs a Boolean, we define the procedure to also work when $f$ outputs a probabilistic value of failure (demonstrated in \cref{sec:prob_valued_sys}).
Throughout, we use the fact that the surrogate model provides a probabilistic interpretation of the failure predictions regardless of the type of system outputs, namely Boolean or probabilistic.

\paragraph{Uncertainty exploration.} To find failures and cover the design space $\mathcal{X}$, we want to explore areas with high uncertainty and high operational likelihood. The first proposed acquisition function searches over the uncertainty provided by the Gaussian process $\hat{\sigma}(\vec{x})$ weighted by the operational model $p(\vec{x})$ to find likely points $\vec{x} \in \mathcal{X}$ with maximal uncertainty:
\begin{equation}
    \vec{x}'_1 = \argmax_{\vec{x} \in \mathcal{X}} \hat{\sigma}(\vec{x})p(\vec{x})^{1/\alpha t} \label{eq:uncertainty_exploration}
\end{equation}
The influence of the operational model is decayed by $1/\alpha t$.
This will ensure that the design space $\mathcal{X}$ is fully explored in the limit \cite{williams2006gaussian}, noting that in practice the limiting factor is the $\mathcal{O}(n^3)$ time for the Gaussian process to fit $n$ data points due to the $n \times n$ matrix inversion \cite{quinonero2005unifying}. Equation \ref{eq:uncertainty_exploration} may also be sampled instead of taking the $\argmax$ where the concentration of the distribution can be controlled by a temperature parameter $\tau$:
\begin{equation}
    \vec{x}'_1 \sim \big(\hat{\sigma}(\vec{x})p(\vec{x})^{1/\alpha t}\big)^{1/\tau} \label{eq:uncertainty_exploration_sampled}    
\end{equation}
This distribution is normalized over the domain to compute the proposal likelihood $q(\vec{x}_i)$ of each sample. The likelihood is used to compute the weight $w(\vec{x}_i) = p(\vec{x}_i) / q(\vec{x}_i)$ for self-normalized importance sampling (see \cref{sec:is}).

\paragraph{Boundary refinement.} To better characterize the areas of all failure regions, we want to refine the known failure boundaries to tighten them as much as possible.
Because our surrogate $\hat{f}(\vec{x})$ is modeled as a logistic function (shown in \cref{eq:sigmoid}), we can take the derivative and get the analytical form as
\begin{equation}
    \mu'(\vec{x}) = \hat{f}(\vec{x})(1 - \hat{f}(\vec{x}))
\end{equation}
where $\mu'(\vec{x})$ is maximal when $\hat{f}(\vec{x}) = 0.5$, thus giving us the failure boundary at the peaks.
Therefore, the second proposed acquisition function selects the point that maximizes the upper confidence of $\mu'$ to refine the failure boundary:
\begin{equation}
    \vec{x}'_2 = \argmax_{\vec{x} \in \mathcal{X}} \bigl(\mu'(\vec{x}) + \lambda\hat{\sigma}(\vec{x})\bigr)p(\vec{x})^{1/\alpha t} \label{eq:boundary_refinement}
\end{equation}
where upper confidence provides an over estimation and the factor parameter is set to $\lambda = 0.1$ in our experiments.
The operational model $p(\vec{x})$ is used to first focus on the failure boundary with high operational likelihood, then decay the emphasis of the likelihood as a function of the current iteration $t$ (here using an inverse decay of $1/t$).
This will first acquire likely points along the boundaries, then refine all of the boundaries because as $t \to \infty$ then $p(\vec{x})^{1/\alpha t} \to 1$.

Similar to \cref{eq:uncertainty_exploration_sampled}, we may also choose to sample along the boundary in \cref{eq:boundary_refinement} with a temperature parameter $\tau$ and compute the weights after normalization:
\begin{equation}
    \vec{x}'_2 \sim \big(\bigl(\mu'(\vec{x}) + \lambda\hat{\sigma}(\vec{x})\bigr)p(\vec{x})^{1/\alpha t}\big)^{1/\tau} \label{eq:boundary_refinement_sampled}
\end{equation}
Sampling these acquisition functions may be advantageous when the black-box system is stochastic, thus making the failure boundaries noisy. Further discussion and analysis of stochastic systems is provided in sections \ref{sec:pomdp_example} and \ref{sec:pomdp_results}.

\paragraph{Failure region sampling.} The optimal importance sampling distribution is
\(
q_\text{opt} \propto f(\vec{x})p(\vec{x}),
\)
which, intuitively, is the distribution of failures (when $f(\vec{x}) = 1$) over the likely region (weighed by $p(\vec{x})$) \cite{kahn1953methods,murphy2012machine,owen2013monte}.
Yet this is exactly what we are trying to estimate and sampling this distribution may require a prohibitive number of evaluations of $f$.
Therefore, the third proposed acquisition function uses the surrogate to get the upper confidence of the failure prediction
\begin{align}
    \hat{h}(\vec{x}) &= \hat{f}(\vec{x}) + \lambda\hat{\sigma}(\vec{x})\\
    \hat{g}(\vec{x}) &= \mathds{1}\bigl\{\hat{h}(\vec{x}) \ge 0.5\bigr\}
\end{align}
and then using the estimated failure region $\hat{g}$ we draw a sample from the distribution
\begin{equation}
    \vec{x}'_3 \sim \hat{g}(\vec{x})p(\vec{x}).
\end{equation}
Here, we use the indicator function $\mathds{1}\{ \cdot \}$, which returns $1$ when the input is \texttt{true} and $0$ otherwise.
Sampling from the approximate failure distribution defined by the surrogate helps to refine likely failure regions to ensure a better estimate of the probability of failure.
If the system $f$ outputs a failure probability value in $[0,1]$ instead of a binary indicator, then we can use this information and sample from the distribution that weights towards higher confidence failures:
\begin{equation}
    \vec{x}'_3 \sim \hat{g}(\vec{x})\hat{h}(\vec{x})p(\vec{x}) \label{eq:frs_pvalue}
\end{equation}

The proposed acquisition functions work under a more restrictive binary system $f: \mathbb{R}^n \to \mathbb{B}$ and a system $f: \mathbb{R}^n \to [0,1]$ that outputs a probabilistic value of failure (which can be interpreted as confidence or stochasticity).
We define \textit{failure region sampling} using the more general distribution in \cref{eq:frs_pvalue} because it works for both types of system outputs.
When a granular measure of system failure is available, it can be used to make a more informative surrogate model.
If only binary failure information is available, developers could simply focus on those failures with high likelihood.
Applying to binary-valued systems is more general and thus the primary focus of this work, but we demonstrate on a probability-valued case in \cref{sec:prob_valued_sys}.
In the case when no failures are predicted (i.e., $\hat{g}(\vec{x}) = 0,\, \forall \vec{x} \in \mathcal{X}$), then we sample directly from $\hat{h}$.

\begin{figure}[b!]
    \centering
    \includesvg[inkscapelatex=false, width=\textwidth]{figures/toys/plot-combined-30-booth-models.svg}
    \caption{Illustrating 90 steps of the \textit{failure search and refinement} acquisition functions.}
    \label{fig:fsar_toy}
\end{figure}

\Cref{alg:fsar} describes the \textit{failure search and refinement} (FSAR) procedure to compute the subsequent points from the three proposed acquisition functions. \Cref{fig:fsar_toy} provides an illustrative example where the probabilistic surrogate model with predicted failures shown in red and the \textit{failure search and refinement} acquisition functions after $T=\num{30}$ iterations are shown (with $N=\num{90}$ data points, showing true observations as red/green squares).
Lighter colors indicate maximums and red circles indicate the next selected point.
The operational likelihood model $p(\vec{x})$ is shown as marginal distribution subplots.
Notice the low uncertainty around the likely failure region and the influence of $p(\vec{x})$ on the boundary refinement; the algorithm first refines the likely boundary, and then refines the entire boundary in the limit. The system under test is an example system shown in \cref{fig:toy_rep}

{\centering
\begin{minipage}{.7\linewidth}\hypertarget{link:fsar}{}
    \linespread{1.0}\selectfont
    \begin{algorithm}[H]
        \caption{Failure search and refinement acquisition functions.} 
        \label{alg:fsar}
        \begin{algorithmic}[1]
        \Function{FailureSearchAndRefinement$(\mathcal{GP}, p, t)$}{}
        \State $\hat{f} \leftarrow \textproc{MeanFunction}(\mathcal{GP})$
        \State $\hat{\sigma} \leftarrow \textproc{StandardDeviationFunction}(\mathcal{GP})$
    
        \State {\color{lightgray}\# 1) uncertainty exploration}
        \State $\vec{x}'_1 \leftarrow \begin{cases}
            \vec{x}' \leftarrow \argmax_{\vec{x} \in \mathcal{X}} \hat{\sigma}(\vec{x})p(\vec{x})^{1/\alpha t} & \text{if } \tau = 0\\
            \vec{x}' \sim \big( \hat{\sigma}(\vec{x})p(\vec{x})^{1/\alpha t} \big)^{1/\tau} & \text{otherwise}
        \end{cases}$
    
        \State {\color{lightgray}\# 2) boundary refinement}
        \State $\mu'(\vec{x}) \leftarrow \hat{f}(\vec{x})(1 - \hat{f}(\vec{x}))$ \GrayComment{compute failure boundaries}
        \State $\vec{x}'_2 \leftarrow \begin{cases}
            \vec{x}' \leftarrow \argmax_{\vec{x} \in \mathcal{X}} \bigl(\mu'(\vec{x}) + \lambda\hat{\sigma}(\vec{x})\bigr)p(\vec{x})^{1/\alpha t} & \text{if } \tau = 0\\
            \vec{x}' \sim \big( \bigl(\mu'(\vec{x}) + \lambda\hat{\sigma}(\vec{x})\bigr)p(\vec{x})^{1/\alpha t} \big)^{1/\tau} & \text{otherwise}
        \end{cases}$
    
        \State {\color{lightgray}\# 3) failure region sampling}
        \State $\hat{h}(\vec{x}) \leftarrow \hat{f}(\vec{x}) + \lambda\hat{\sigma}(x)$ \GrayComment{compute upper confidence bound}
        \State $\hat{g}(\vec{x}) \leftarrow \mathds{1}\bigl\{\hat{h}(\vec{x}) \ge 0.5\bigr\}$ \GrayComment{compute failure regions}
        \State $\vec{x}'_3 \leftarrow \begin{cases}
            \vec{x}' \sim \hat{h}(\vec{x}) & \text{if } \hat{g}(\vec{x}) = 0,\, \forall \vec{x} \in \mathcal{X} \\
            \vec{x}' \sim \hat{g}(\vec{x})\hat{h}(\vec{x})p(\vec{x}) & \text{otherwise}
        \end{cases}$

        \State $\vec{w}' \leftarrow \textproc{ComputeWeights}(\vec{x}'_1, \vec{x}'_2, \vec{x}'_3)$
        \State \Return $\{\vec{x}'_1, \vec{x}'_2, \vec{x}'_3\}, \vec{w}'$
        \EndFunction
        \end{algorithmic}
    \end{algorithm}
\end{minipage}
\par}

\subsection{Importance Sampling Estimate of Failure Probability}\label{sec:is}

To compute an efficient and unbiased estimate of the probability of failure, we use \textit{importance sampling} \cite{owen2013monte}.
Probability estimation can be defined as computing the expectation of the Boolean-valued function $f$ over the \textit{target/nominal distribution} $p$ (what we call the \textit{operational likelihood model} in this work) as
\begin{equation}\label{eq:expectation}
    \mathbb{P}[f(\vec{x})] = \operatorname*{\mathbb{E}}_{\vec{x} \sim p}[f(\vec{x})] = \int_\mathcal{X} p(\vec{x})f(\vec{x}) \ dx.
\end{equation}
In general, the expectation of the indicator function of an event $A$, denoted $\mathds{1}\{A\}$, is equal to the probability of that event occurring $\mathbb{E}[\mathds{1}\{A\}] = \mathbb{P}[A]$.
In our problem, we define $f: \mathbb{R}^n \to \mathbb{B}$ as a Boolean-valued function for convenience.
Nevertheless, the following work could easily be extended to a real-valued function $v: \mathbb{R}^n \to \mathbb{R}$ where failures are defined by violating some safety threshold $c$, i.e., $f(\vec{x}) = \mathds{1}\{v(\vec{x}) \ge c\}$.
Now to approximate the expectation---and therefore the probability of failure---we can use $n$ samples from $p$:
\begin{equation}
    \operatorname*{\mathbb{E}}_{\vec{x} \sim p}[f(\vec{x})] \approx \frac{1}{n} \sum_{i=1}^n f(\vec{x}_i)
\end{equation}
If failures are rare under the distribution $p$ (i.e., $f(\vec{x}_i)$ is rarely equal to $1$ when $\vec{x}_i \sim p$), then we may need an extremely large number of samples from $p$ to get an accurate estimate.
But this would require prohibitively many system evaluations of $f$. 
Instead, importance sampling states that we can sample from some other distribution $q$, called the \textit{proposal distribution}, and re-weight the outputs of $f$ based on the \textit{likelihood ratio} $p(\mathbf{x})/q(\mathbf{x})$ \cite{owen2013monte}:
\begin{align}
    \operatorname*{\mathbb{E}}_{\vec{x} \sim p}[f(\vec{x})] &= \operatorname*{\mathbb{E}}_{\vec{x} \sim q}\left[\frac{p(\vec{x})}{q(\vec{x})}f(\vec{x})\right] \\
    &\approx \frac{1}{n} \sum_{i=1}^n \frac{p(\vec{x}_i)}{q(\vec{x}_i)}f(\vec{x}_i)
\end{align}
Now we can use samples from $q$ to approximate the probability over $p$.
However, selecting an effective proposal distribution can be challenging (see \citet{bugallo2017adaptive}).
Based on whether the black-box system is stochastic or deterministic, we can employ different importance sampling methods to estimate the proposal distribution.

\paragraph{Discrete proposal.}
In the case of deterministic systems, where every input $\vec{x}$ maps deterministically to an output $y$, we could use a uniform proposal over the design space $q = \mathcal{U}_\mathcal{X}$ and replace the expensive function calls to $f$ with inexpensive evaluations of the surrogate $\hat{f}$ using orders of magnitude more samples.
We let $\hat{g}(\vec{x}) = \mathds{1}\{\hat{f}(\vec{x}) \ge 0.5\}$ to indicate failures predicted by the surrogate model.
Thus our problem gets simplified to estimating
\begin{equation}
    \hat{p}_\text{fail} = \operatorname*{\mathbb{E}}_{\vec{x} \sim p}[\hat{g}(\vec{x})] = \operatorname*{\mathbb{E}}_{\vec{x} \sim q}\left[\frac{p(\vec{x})}{q(\vec{x})}\hat{g}(\vec{x})\right] \approx \frac{1}{n} \sum_{i=1}^n \frac{p(\vec{x}_i)}{q(\vec{x}_i)}\hat{g}(\vec{x}_i).\label{eq:pfail}
\end{equation}
Using a uniform distribution can induce variance in the estimate \cite{bishop2006pattern}, therefore an even further simplification is to use a discretized set of $n$ points $\bar{\mathcal{X}}$ over the range $\mathcal{X}$ as our proposal.
We assigned equal likelihood to each point $\vec{x}_i \in \bar{\mathcal{X}}$, namely $q(\vec{x}_i) = 1/n\sum_{j=1}^n p(\vec{x}_j)$.
Then \cref{eq:pfail} becomes
\begin{equation}
    \hat{p}_\text{fail} \approx \frac{1}{n} \sum_{i=1}^n \frac{p(\vec{x}_i)}{1/n\sum_{i=1}^n p(\vec{x}_i)}\hat{g}(\vec{x}_i) = \frac{\sum_{i=1}^n p(\vec{x}_i)\hat{g}(\vec{x}_i)}{\sum_{i=1}^n p(\vec{x}_i)} = \frac{\vec{w}^\top\hat{\vec{y}}}{\sum_{i=1}^n w_i}
\end{equation}
where $\vec{w} = [p(\vec{x}_1), \ldots, p(\vec{x}_n)]$ and $\hat{\vec{y}} = [\hat{g}(\vec{x}_1), \ldots, \hat{g}(\vec{x}_n)]$ for $\vec{x}_i \in \bar{\mathcal{X}}$.
Here, we are using \textit{likelihood weighting}, which is a special case of importance sampling \cite{bishop2006pattern,murphy2012machine}.
Using a discrete set of points as the proposal distribution has lower variance than sampling the uniform space, but may not scale well to higher dimensions.
For this paper, we use a simplified $500 \times 500$ discrete grid as the proposal for two-dimensional systems.
To incorporate better proposal distributions when scaling to higher dimensions, see \citet{bugallo2017adaptive} for adaptive importance sampling methods.

\paragraph{Self-normalizing importance sampling.}
To address the case of stochastic systems, where the output of the system is a random variable, we use the surrogate model to guide the search based on the FSAR acquisition functions and collect the 
weights $w(\vec{x})$ from the sampled points of the acquisition functions to estimate the failure probability as
\begin{align}
    \hat{p}_\text{fail} = \frac{\sum_{i=1}^n w(\vec{x}_i) y_i}{\sum_{i=1}^n w(\vec{x}_i)}
\end{align}
which is the \textit{self-normalized importance sampling estimate} and will equal the true target value in the limit \cite{owen2013monte}.
The variance estimate is computed as
\begin{equation}
    \widehat{\operatorname{Var}}(\hat{p}_\text{fail}) = \sum_{i=1}^n \bar{w}(\vec{x}_i)^2 (y_i - \hat{p}_\text{fail})^2    
\end{equation}
where $\bar{w}(\vec{x}_i) = w(\vec{x}_i) / \sum_{j=1}^n w(\vec{x}_j)$ and the $99\%$ confidence interval is $\hat{p}_\text{fail} \pm 2.58 \sqrt{\widehat{\operatorname{Var}}(\hat{p}_\text{fail})}$.
Because failures are stochastic, applying the self-normalized weights to the true system outputs $y_i \in \vec{Y}$ means we can estimate the failure probability over observed failures and compute proposal weights for only the observed points using the surrogate, thus avoiding computing the surrogate over a large discrete grid proposal.

\begin{algorithm}[t!]
    \caption{Bayesian safety validation algorithm.} 
    \label{alg:bsv}
    \begin{algorithmic}[1]
    \Function{BayesianSafetyValidation$(f, p, T)$}{}
    \State $\mathcal{GP} \leftarrow \textproc{InitializeGaussianProcess}(m, k)$
    \State $\vec{X}, \vec{Y}, \vec{W} \leftarrow \emptyset, \emptyset, \emptyset$
    \For {$t \leftarrow 1$ \textbf{to} $T$}
        \State $\vec{X}', \vec{W}' \leftarrow \hyperlink{link:fsar}{\textproc{FailureSearchAndRefinement}}(\mathcal{GP}, p, t)$ \GrayComment{select new design points (\cref{alg:fsar})} \label{line:subsequent_points}
        \State $\vec{Y}' \leftarrow \{f(\vec{x}') \mid \vec{x}' \in \vec{X}'\}$ \GrayComment{evaluate true system $f$ across design points}
        \State $\vec{X}, \vec{Y}, \vec{W} \leftarrow \vec{X} \cup \vec{X}',\, \vec{Y} \cup \vec{Y}',\, \vec{W} \cup \vec{W}'$ \GrayComment{append to input, output, and weight sets}

        \State $\mathcal{GP} \leftarrow \textproc{Fit}(\mathcal{GP}, \vec{X}, \vec{Y})$ \GrayComment{refit surrogate model over all points conditioned on observations}
    \EndFor
    \State $\vec{X}_\text{fail} \leftarrow \big\{\vec{x}_i \mid \vec{x}_i \in \vec{X},\, y_i \in \vec{Y},\, \mathds{1}\{y_i\}\big\}$ \GrayComment{(1) falsification (set of all true failures)} \label{line:falsification}
    \State $\vec{x}^* \leftarrow \argmax_{\vec{x}_i \in \vec{X}, y_i \in \vec{Y}} p(\vec{x}_i)\mathds{1}\{y_i\}$ \GrayComment{(2) most-likely failure} \label{line:mlfa}
    \State $\hat{p}_\text{fail} \leftarrow \begin{cases}
        \frac{1}{n} \sum_{i=1}^n \frac{p(\vec{x}_i)}{q(\vec{x}_i)} \mathds{1}\left\{\hat{f}(\vec{x}_i) \ge 0.5 \right\} & \text{if deterministic} \\
        \left(\sum_{i=1}^n \vec{W}_i \vec{Y}_i\right) / \sum_{i=1}^n \vec{W}_i & \text{if stochastic}
    \end{cases}$ \GrayComment{(3) failure probability estimate} \label{line:fpe}
    \State \Return $\vec{X}_\text{fail}, \vec{x}^*, \hat{p}_\text{fail}$ \GrayComment{return all three safety validation tasks}
    \EndFunction
    \end{algorithmic}
\end{algorithm}

\begin{figure}[t!]
    \centering
    \resizebox{0.96\textwidth}{!}{
        \input{figures/bayesian-safety-validation-diagram}
    }
    \caption{The proposed \textit{Bayesian safety validation} (BSV) algorithm used for all three safety validation tasks.}
    \label{fig:algorithm}
\end{figure}

\subsection{Proposed Algorithm: Bayesian Safety Validation}

The proposed \textit{Bayesian safety validation} (BSV) algorithm takes as input the black-box system $f: \mathbb{R}^n \to \mathbb{B}$ or $f: \mathbb{R}^n \to [0,1]$ and an operational likelihood model $p: \mathbb{R}^n \to \mathbb{R}_{\ge 0}$ for inputs $\vec{x} \in \mathbb{R}^n$ of some parametric space, and iteratively refits a probabilistic surrogate model given selected points from the FSAR acquisition functions (\cref{sec:problem}).
The first acquisition, \textit{uncertainty exploration}, explores areas with high uncertainty to provide coverage and search for failure regions.
The next acquisition, \textit{boundary refinement}, selects operationally likely points that refine the failure boundaries to better characterize likely failure regions (and includes a decaying weighted operational likelihood to refine all failure boundaries in the limit).
The final acquisition, \textit{failure region sampling}, is based on the theoretically optimal $q$-proposal distribution \cite{kahn1953methods} and will sample from the likely failure regions to ensure a better estimate of the probability of failure.
After the algorithm runs for $T$ iterations, a total of $3T$ sampled points were used to fit the surrogate model.
The three safety validation tasks are then computed (lines \ref{line:falsification}--\ref{line:fpe}).
Falsification and most-likely failure analysis use only the true observations $y_i \in \vec{Y}$ and actual inputs $\vec{x}_i \in \vec{X}$ to find those inputs that led to failures and the most-likely failure, respectively.
Then the final surrogate model or the weights are used to efficiently compute an importance sampling estimate of the failure probability.
\Cref{alg:bsv} describes the BSV algorithm, and \cref{fig:algorithm} illustrates the process.

\section{Experiments and Results}\label{sec:experiments}

To test the effectiveness of BSV, we ran experiments across several different example systems and a real-world case study using a prototype neural network-based runway detection system.
We split the experiments into two sections: 1) comparison against existing methods for rare-event simulation (this tests the full Bayesian safety validation algorithm), and 2) comparison of the Gaussian process-based approach with different sampling/selection methods (this tests the failure search and refinement acquisition functions).
We ran an ablation study to empirically show the influence of each acquisition function on the performance of the safety validation tasks.
We demonstrate the algorithm on a stochastic system and a system that outputs a probabilistic value of failure (instead of strictly binary) to show that BSV is applicable in the less restrictive problem case.
Lastly, we report results on the runway detection system as a real-world case study.

\begin{figure}[t!]
     \centering
     \begin{subfigure}[b]{0.31\textwidth}
         \centering
         \includesvg[inkscapelatex=false, width=\textwidth]{figures/toys/truth_booth_models_joint.svg}
         \caption{Representative problem (Booth's).}
         \label{fig:toy_rep}
     \end{subfigure}
     \hfill
     \begin{subfigure}[b]{0.31\textwidth}
         \centering
         \includesvg[inkscapelatex=false, width=\textwidth]{figures/toys/truth_squares_models_joint.svg}
         \caption{Two square failure modes.}
         \label{fig:toy_squares}
     \end{subfigure}
     \hfill
     \begin{subfigure}[b]{0.31\textwidth}
         \centering
         \includesvg[inkscapelatex=false, width=\textwidth]{figures/toys/truth_himmelblau_models.svg}
         \caption{Mixture model, three failure modes.}
         \label{fig:toy_mixture}
     \end{subfigure}
        \caption{Failure regions and operational models for the three test problems.}
        \label{fig:toys}
\end{figure}

\subsection{Simplified Test Problems}\label{sec:toys}
Three example toy problems with access to the true value of $p_\text{fail}$ were used for testing.
The first problem (called \textsc{Representative}) was chosen based on the observed shape of the failure region of the runway detection system, which is our primary case study and a system which we do not have access to the true failure probability.
The representative toy problem is modeled using Booth's function \cite{optbook} $f(\vec{x}) = (x_1 + 2x_2 - 7)^2 + (2x_1 + x_2 - 5)^2 \le 200$, thresholded to make this a binary function.
We define the operational parameters to be over the range $[-10, 5]$ for both $x_1$ and $x_2$ and set the operational likelihood model as $x_1 \sim \mathcal{N}_\text{trunc}(-10, 1.5; [-10, 5])$ and $x_2 \sim \mathcal{N}(-2.5, 1)$ where $\mathcal{N}_\text{trunc}(\mu, \sigma; [a, b])$ is the normal distribution truncated between $[a,b]$.
The second toy problem (called \textsc{Squares}) has two, square, disjoint failure regions to test the exploration of BSV and refinement of rigid and disjoint failure boundaries.
The operational parameters are over the range $[0, 10]$ for both $x_1$ and $x_2$, each with the operational likelihood model of $\mathcal{N}(5, 1)$.
The third toy problem (called \textsc{Mixture}) has three, smooth, disjoint failure regions and is designed to test the failure region refinement characteristic of BSV using a multimodal operational model.
The operational range is over $[-6, 6]$ with identical Gaussian mixture models that have two equal components of $\mathcal{N}_\text{trunc}(2,1; [-6,6])$ and $\mathcal{N}_\text{trunc}(-2,1; [-6,6])$.
Similar to the representative example, we define this last problem as the thresholded Himmelblau function \cite{himmelblau1972applied}: $f(\vec{x}) = (x_1^2 + x_2 - 11)^2 + (x_1 + x_2^2 - 7)^2 \le 15$.
The test problems and their operational models are shown in \cref{fig:toys} where failure regions (in red) are shown for the test problems.
Operational models are illustrated as subplots/contours and the true system is shown above the surrogate model failure classification fit with $\num{999}$ samples using BSV.

\subsection{Stochastic Sequential Decision Making System}\label{sec:pomdp_example}
BSV also works in cases where system failures are stochastic, meaning that the same input could lead to a different failure outcome.
To test a stochastic system, we benchmark on the \textit{LightDark} sequential decision making problem \cite{platt2010belief}, modeled as a \textit{partially observable Markov decision process} (POMDP).
The LightDark POMDP involves long-horizon localization, allowing the agent to move \texttt{up} or \texttt{down} by one to find the light region at $y=10$ and receives noisy observations of its true position as a function of the distance to the light region.
The agent will receive a large reward of $100$ when executing the \texttt{stop} action at $\pm 1$ of the origin.
If the agent stops outside the origin, then it receives a large penalty of $-100$.
We test BSV under two cases: 1) a rare failure event when the agent never executes a \texttt{stop} action, and 2) a non-rare failure event when either the agent never executes a \texttt{stop} action or the \texttt{stop} action is executed outside the origin.
The problem is one-dimensional and we use the initial state distribution of $\mathcal{N}(2, 3)$ as the operational model.

\subsection{Neural Network-based Runway Detection System}

As a real-world case of a safety-critical subsystem, we chose a common application in autonomous flight: runway detection (RWD) using neural networks.
Synthetic images were generated using the flight simulator X-Plane \cite{xplane}, sampling over different parameters of the final approach to land (e.g., glide slope angle and distance to runway).
We search over this parametric space instead of dealing directly in pixel or image space.
We use an operational model of likely glide slope angles with a small standard deviation, namely $\alpha \sim \mathcal{N}(3, 0.5)$, and a model that increases the likelihood of requiring a detection as the distance to the runway decreases, namely $d \sim \mathcal{N}_\text{trunc}(0, 1; [0, 4])$.
The parametric space is continuous in glide slope angle $\alpha \in [1,7]$ degrees and distance to runway $d \in [0.1,4]$ nmi.
These models can be learned from historical flight data for more accurate estimates of the failure probability.

\begin{figure}[t!]
     \centering
     \begin{subfigure}[b]{0.3\textwidth}
         \centering
         \includegraphics[width=\textwidth]{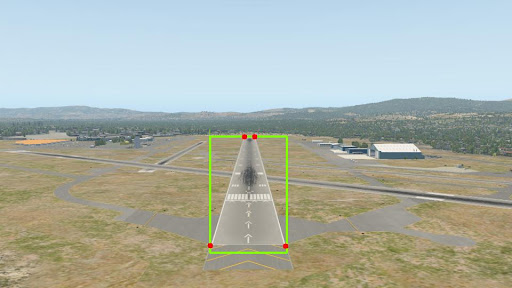}
         \caption{Nominal runway conditions.}
         \label{fig:radnet_nominal}
     \end{subfigure}
     \hfill
     \begin{subfigure}[b]{0.3\textwidth}
         \centering
         \includegraphics[width=\textwidth]{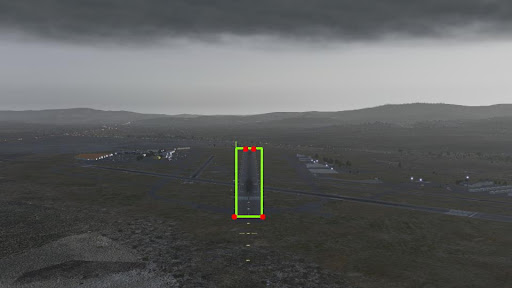}
         \caption{Low-likely weather conditions.}
         \label{fig:radnet_weather}
     \end{subfigure}
     \hfill
     \begin{subfigure}[b]{0.3\textwidth}
         \centering
         \includegraphics[width=\textwidth]{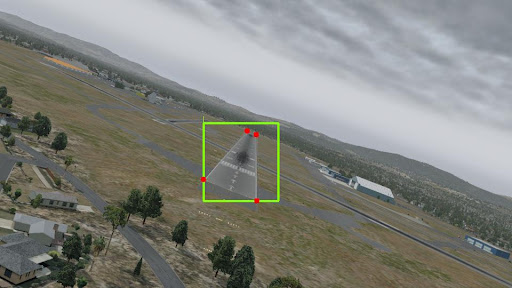}
         \caption{Potentially dangerous to fly.}
         \label{fig:radnet_dangerous}
     \end{subfigure}
        \caption{Applying the neural network runway detector to simulated runway conditions in X-Plane.}
        \label{fig:radnet_xplane}
\end{figure}

Treated as a black box, the runway detector is a convolutional neural network that processes runway images from a front-facing RGB camera. The network predicts the runway corners and the runway bounding box and is intended to be used as a subsystem to provide position estimates during autonomous landing \cite{durand2023formal}.
\Cref{fig:radnet_xplane} illustrates several example images with detected runway corners and bounding boxes.
A failure is defined as a misdetection (i.e., a false negative). Since the system is designed to be active only during the landing phase, we condition on the aircraft being on the approach.
The use of a simulator means the runway detection system can be stressed outside the normal flight envelope to better characterize the full range of system failures, with a potentially dangerous-to-fly example shown in \cref{fig:radnet_dangerous}.

\subsection{Safety Validation Metrics}
We define several metrics to measure the performance across the three safety validation tasks (see \cref{sec:safety}).

\paragraph{Falsification metrics.} The total number of failure cases, or more generally, the proportion of all system evaluations that resulted in failures is the primary metric used to assess falsification (sometimes called the failure rate):
\begin{equation}
    R_\text{fail} = \frac{\text{number of failures}}{\text{total number of evaluations}} = \frac{|\vec{X}_\text{fail}|}{|\vec{X}|}
\end{equation}

\paragraph{Most-likely failure analysis metrics.} The goal of most-likely failure analysis, as the name suggests, is to determine the failure with maximum operational likelihood. A natural way to assess the relative performance of this task against baselines is to compare the likelihood of the determined most-likely failure:
\begin{equation}
    \mathcal{L}^* = \max_{\vec{x}_i \in \vec{X}, y_i \in \vec{Y}} p(\vec{x}_i)\mathds{1}\{y_i\}
\end{equation}

\paragraph{Failure probability estimation metrics.} Because probability of failure estimation is the primary objective of this work---capturing all three safety validation tasks---we look at the performance across several different metrics.
When we have access to the true $p_\text{fail}$ (e.g., in the toy examples), then we can measure the relative error in the estimated $\hat{p}_\text{fail}$:
\begin{equation}
    \hat{\Delta}_\text{fail} = \frac{|p_\text{fail} - \hat{p}_\text{fail}|}{p_\text{fail}} \label{eq:pfail_error}
\end{equation}
Measured as a proportion, relative error can be interpreted as the percent difference in the estimate and makes it easier to compare performance across problems.
We also analyze the failure likelihood distribution $\{\log p(\vec{x}_i)\}_{\vec{x}_i \in \vec{X}_\text{fail}}$, where distributions with higher likelihood are preferred as they cover more relevant example failures.

\paragraph{Coverage of design space.} To measure the coverage of the design space, an average dispersion coverage metric has been used in the context of safety validation to estimate how well the sampled points cover the input space \cite{esposito2005adaptive, corso2021survey}:
\begin{equation}
    C_\text{input}(\vec{X}) = 1 - \frac{1}{\delta} \sum_{j=1}^n \frac{\min(d_j(\vec{X}), \delta)}{n}
\end{equation}
where $C_\text{input}(\vec{X}) \in [0,1]$ and the metric is defined over a grid of $n$ points, separated by $\delta$.
The distance function $d_j(\vec{X})$ is defined as the minimum distance between the $j$th point in the grid to a point in $\vec{X}$ \cite{esposito2005adaptive, corso2021survey}.

When ground truth is available, we are also interested in the characterization of the failure and non-failure regions over the entire domain as predicted by the surrogate model. We define $C_\text{output} \in [0,1]$ as the proportion of the output space that the surrogate and the true system agree upon.
This can be interpreted as the surrogate classification accuracy.

\begin{table*}[b!]
\caption{Comparison against other sampling/selection methods.}\label{tab:ablation_sampling}
\vspace*{-5mm}
\begin{center}
\begin{threeparttable}
    \begin{small}
    \begin{tabular}{@{}clrrrrr@{}}
    \toprule
        Example Problem & Selection Method & $R_\text{fail} \uparrow$ & $\mathcal{L}^* \uparrow$ & $\hat{\Delta}_\text{fail} \downarrow$ & $C_\text{input} \uparrow$ & $C_\text{output} \uparrow$\\
        \midrule
        \textsc{Representative} %
        & Latin hypercube sampling      &  $0.332$                 &  $\num{2.49e-8}$           &  $\num{0.44777}$             &  $0.682$                 &  $0.9922$  \\
        \multirow{4}{*}{\makecell[l]{\includesvg[width=0.5in]{figures/toys/surrogate_hard_booth_small.svg}}} %
        & Sobol sequence sampling       &  $0.335$                 &  $\num{4.98e-8}$           &  $\num{0.51179}$             &  $0.719$                 &  $0.9912$  \\
        & Discrete grid selection       &  $0.336$                 &  $\num{5.62e-8}$           &  $\num{0.58225}$             &  \bfseries$\num{0.774}$  &  $0.9933$  \\
        & Uniform sampling              &  $0.342$                 &  $\num{4.02e-8}$           &  $\num{0.25978}$             &  $0.673$                 &  $0.9919$  \\
        & Failure search and refinement (Ours) &  \bfseries$\num{0.585}$  &  \bfseries$\num{5.78e-8}$  &  \bfseries$\num{0.00667}$    &  $0.638$                 &  \bfseries$\num{0.9998}$  \\
        \midrule
        \textsc{Squares} %
        & Latin hypercube sampling      &  $0.0525$                 &  $0.00192$                 &  $\num{0.24486}$            &  $0.682$                 &  $0.9907$  \\
        \multirow{4}{*}{\makecell[l]{\includesvg[width=0.5in]{figures/toys/surrogate_hard_squares_small.svg}}} %
        & Sobol sequence sampling       &  $0.0525$                 &  $0.00142$                 &  $\num{0.27050}$            &  $0.719$                 &  $0.9935$  \\
        & Discrete grid selection       &  $0.0439$                 &  $0.00196$                 &  $\num{0.26473}$            &  \bfseries$\num{0.774}$  &  $0.9894$  \\
        & Uniform sampling              &  $0.0532$                 &  $0.00086$                 &  $\num{0.17017}$            &  $0.673$                 &  $0.9909$  \\
        & Failure search and refinement (Ours) &  \bfseries$\num{0.4800}$  &  \bfseries$\num{0.00253}$  &  \bfseries$\num{0.00727}$   &  $0.643$                 &  \bfseries$\num{1.0}$  \\
        \midrule
        \textsc{Mixture} %
        & Latin hypercube sampling      &  $0.0441$                  &  $0.0349$                  &  $\num{0.10469}$           &  $0.682$                 &  $0.9864$  \\
        \multirow{4}{*}{\makecell[l]{\includesvg[width=0.5in]{figures/toys/surrogate_hard_himmelblau_small.svg}}} %
        & Sobol sequence sampling       &  $0.0505$                  &  $0.0369$                  &  $\num{0.00787}$           &  $0.719$                 &  $0.9881$  \\
        & Discrete grid selection       &  $0.0479$                  &  $0.0345$                  &  $\num{0.00279}$           &  \bfseries$\num{0.774}$  &  $0.9900$  \\
        & Uniform sampling              &  $0.0576$                  &  $0.0360$                  &  $\num{0.04919}$           &  $0.673$                 &  $0.9871$  \\
        & Failure search and refinement (Ours) &  \bfseries$\num{0.4640}$   &  \bfseries$\num{0.0383}$   &  \bfseries$\num{0.00124}$  &  $0.663$                 &  \bfseries$\num{0.9984}$  \\
        \bottomrule
    \end{tabular}
    \end{small}
\end{threeparttable}
\end{center}
\end{table*}

\subsection{Baseline Methods}
We compare \textsc{BayesianSafetyValidation} (\cref{alg:bsv}) against standard Monte Carlo (MC) sampling and population Monte Carlo (PMC) \cite{cappe2004population} with self-normalized importance sampling \cite{owen2013monte}.
PMC requires an initial adaptive proposal $q_\text{PMC}$, which we set to be equal to the operational likelihood model for each of the example problems.
The experiments were run for $T_\text{max} = 100$ iterations using $N_q = 50$ samples per iteration and ran across $3$ seeds.
This results in $N_qT_\text{max}(T_\text{max} + 1)/2 = 252{,}500$ total samples per seed.
Because sampling-based methods like MC and PMC tend to require many samples to adequately estimate the rare-event \cite{de2005tutorial}, we only test these methods on the example toy problems.
Motivated by sample efficiency, we focus our comparison on the relative error in the estimated probability of failure as a function of the number of samples, defined in \cref{eq:pfail_error}.
\Cref{fig:pmc} shows the error curves over the number of samples, which is equivalent to the number of system evaluations.
Note that MC fails to estimate anything in the allotted number of samples in the representative example.
BSV outperforms both MC and PMC in reducing the error in the probability of failure estimate using several orders of magnitude fewer samples; converging before $1000$ samples in each case and closer to $100$ samples in the first two problems.
Results also indicate that BSV has lower variance compared to MC and PMC, which can partially be explained by the fact that two of the three acquisition functions take deterministic maximums and only one, the failure region sampling acquisition, samples predicted failure points stochastically.

To test the proposed \textsc{FailureSearchAndRefinement} procedure (\cref{alg:fsar}),
we use the same GP fitting technique as in \cref{alg:bsv} but replace the selection process in line \ref{line:subsequent_points} with several baseline methods.
The baselines we use are Latin hypercube sampling (LHS) \cite{mckay1979latin}, Sobol sequence selection \cite{sobol1967distribution}, discrete grid selection, and uniform sampling.
Each technique is defined over the entire operational domain.
Importantly, we note that all methods fit the selected points to the same initial GP and use the same importance sampling procedure defined in \cref{alg:bsv}.
The FSAR approach is the only method that uses incremental information to optimize the subsequent points.
\Cref{fig:sampling} illustrates the relative error in the estimate when running BSV for $T=333$ iterations ($N=999$ system evaluations), run over 3 seeds with shaded regions reporting standard deviation (noting that Sobol and discrete do not use stochasticity).
Exponential smoothing is applied to the curves with the raw values as thin lines of the same color.
Using FSAR for acquiring subsequent points outperforms the baselines by orders of magnitude in the first two problems, and is comparable to Sobol sequence selection in the third problem but with more stability in the estimate.
\Cref{tab:ablation_sampling} reports the quantitative results from the baseline experiments.
FSAR achieves the best performance across the various safety validation metrics and comparable input coverage relative to the baselines.

\begin{figure}[t!]
    \begin{center}
     \begin{subfigure}[t]{0.49\textwidth}
        \centering
        \resizebox{0.8271\textwidth}{!}{%
            \input{figures/pmc/pmc_booth}
        }
        \resizebox{0.8271\textwidth}{!}{%
            \input{figures/pmc/pmc_squares}
        }
        \resizebox{0.8271\textwidth}{!}{%
            \input{figures/pmc/pmc_himmelblau}
        }
        \caption{Estimator comparison (with a log-scale horizontal axis).}
        \label{fig:pmc}
     \end{subfigure}
     \hfill
     \begin{subfigure}[t]{0.49\textwidth}
        \centering
        \resizebox{0.8776\textwidth}{!}{%
            \input{figures/toys/baseline_booth}
        }
        \resizebox{0.8776\textwidth}{!}{%
            \input{figures/toys/baseline_squares}
        }
        \resizebox{0.8776\textwidth}{!}{%
            \input{figures/toys/baseline_himmelblau}
        }
        \caption{Fitting the same GP using different sampling schemes.}
        \label{fig:sampling}
     \end{subfigure}
     \caption{Test problem baseline results. Shaded regions show standard error in (a) and standard deviation in (b).}
    \label{fig:performance}
    \end{center}
\end{figure}

\begin{figure}[b!]
    \centering
    \includesvg[pretex=\footnotesize, inkscapelatex=true, width=\textwidth]{figures/toys/aircraft_baselines.svg}
    \caption{Baseline comparison on a complex failure region shape fitting the same GP using $\num{300}$ selected points.}
    \label{fig:aircraft}
\end{figure}

\paragraph{Complex failure region.} As a visual example, we test the baseline sampling algorithms against BSV on a more complex failure region shape in \cref{fig:aircraft} (where the white circles indicate the selected points for each algorithm).
Using a uniform operational likelihood model, BSV is run using only the \textit{boundary refinement} acquisition function which, by design, will explore the uncertainty when no failure boundaries are available to refine.
In relatively few samples given a limited budget, BSV is able to fit to the complex failure region by focusing its search on failure boundary refinement.

\subsection{Ablation Study}
To empirically test the importance of all three acquisition functions, we perform an ablation study on the disjoint squares problem to determine the effect of the combinations of acquisition functions across the safety validation metrics.
The \textsc{Squares} example problem was chosen based on having two disjoint failure regions with precise boundaries; one region which is less likely than the other.
Thus this problem requires careful balance between boundary refinement and deliberate exploration for multiple potential failure regions.
Each ablation was run with $90$ samples over $5$ RNG seeds for a fair comparison (i.e., when using all three acquisitions, each one gets a third of the budget).
Results in \cref{tab:ablation} indicate that the individual acquisitions perform well on the single metric they were designed for (i.e., \textit{exploration} covers the input space, \textit{failure sampling} has the highest failure rate, yet \textit{boundary refinement} requires exploration in order to avoid exploiting a single failure mode). 
Using all three acquisition functions balances between the safety validation metrics and achieves the smallest error in the probability of failure estimate $\hat{\Delta}_\text{fail}$ while also finding a failure with the highest relative likelihood.
In tables \ref{tab:ablation_sampling} and \ref{tab:ablation}, arrows indicate whether the given metric is better to be high $(\uparrow)$ or low $(\downarrow)$.

\begin{table*}[!t]
\caption{Ablation study of the effect of the three \textit{failure search and refinement} acquisition functions.}\label{tab:ablation}
\vspace*{-5mm}
\begin{center}
\begin{threeparttable}
    \begin{small}
    \begin{tabular}{@{}lrrrrr@{}}
    \toprule
        Acquisition(s) & $R_\text{fail} \uparrow$ & $\mathcal{L}^* \uparrow$ & $\hat{\Delta}_\text{fail} \downarrow$ & $C_\text{input} \uparrow$ & $C_\text{output} \uparrow$\\
        \midrule
        $[1]$ exploration                                               &  $\num{0.044}$               &  $\num{0.00029}$             &  $\num{0.04382}$              &  \bfseries$\num{0.233}$       &  $\num{0.9795}$  \\
        $[2]$ boundary refinement                                       &  $\num{0.411}$               &  $\num{0.00025}$             &  $\num{0.93670}$              &  $\num{0.056}$                &  $\num{0.9598}$  \\
        $[3]$ failure sampling                                          &  \bfseries$\num{0.707}$      &  $\num{0.00186}$             &  $\num{0.21682}$              &  $\num{0.066}$                &  $\num{0.9604}$  \\
        $[1,2]$ exploration + boundary refinement                       &  $\num{0.189}$               &  $\num{0.00205}$             &  $\num{0.18483}$              &  $\num{0.159}$                &  \bfseries$\num{0.9801}$  \\
        $[2,3]$ boundary refinement + failure sampling                  &  $\num{0.436}$               &  $\num{0.00197}$             &  $\num{0.24817}$              &  $\num{0.087}$                &  $\num{0.9647}$  \\
        $[1,3]$ exploration + failure sampling                          &  $\num{0.318}$               &  $\num{0.00171}$             &  $\num{0.10057}$              &  $\num{0.163}$                &  $\num{0.9761}$  \\
        $[1,2,3]$ exploration + boundary refinement + failure sampling  &  $\num{0.298}$               &  \bfseries$\num{0.00243}$    &  \bfseries$\num{0.03553}$     &  $\num{0.138}$                &  $\num{0.9777}$  \\
        \bottomrule
    \end{tabular}
    \end{small}
\end{threeparttable}
\end{center}
\end{table*}

\subsection{Stochastic System Results}\label{sec:pomdp_results}
\begin{figure}[ht!]
     \centering
     \begin{subfigure}[t]{0.49\textwidth}
        \centering
        \includegraphics[width=\textwidth]{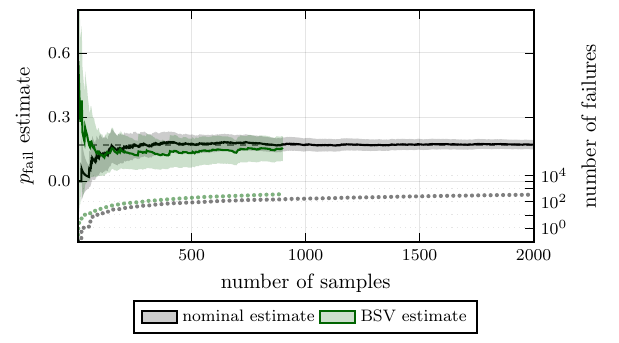}
        \caption{Non-rare stochastic failure.}
        \label{fig:pomdp_nonrare}
     \end{subfigure}
     \hfill
     \begin{subfigure}[t]{0.49\textwidth}
        \centering
        \includegraphics[width=\textwidth]{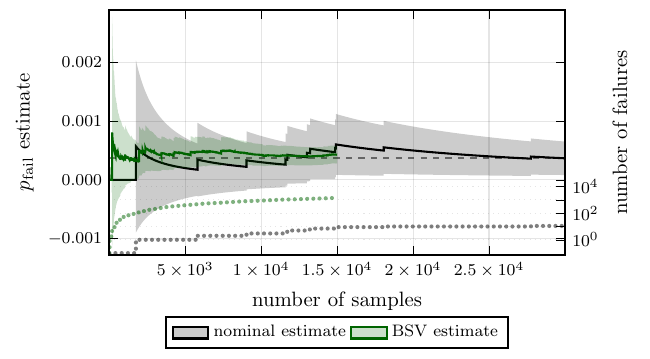}
        \caption{Rare stochastic failure.}
        \label{fig:pomdp_rare}
     \end{subfigure}
     \caption{Failure probability estimate for the stochastic system, comparing nominal sampling vs. BSV.}
    \label{fig:pomdp}
\end{figure}

\Cref{fig:pomdp} shows the failure probability estimation results of the stochastic LightDark POMDP when comparing BSV to nominal estimation, where nominal estimation samples the operational model directly.
For the non-rare case in \cref{fig:pomdp_nonrare}, BSV is comparable to nominal yet finds more failures ($41\%$ failure rate for BSV and $15\%$ for nominal).
The non-rare failure probability estimate shown in the dashed horizontal line is computed after running nominal estimation for $5000$ iterations and comes out to $\hat{p}_\text{fail}=0.15$.
For the rare case in \cref{fig:pomdp_rare}, BSV finds failures early in the search and has a stable failure probability estimate with lower variance using fewer samples than nominal.
Notably, BSV finds orders of magnitude more failures than nominal with a failure rate of about $9\%$ compared to $0.041\%$.
The dashed horizontal line is the nominal estimate computed over $100{,}000$ samples and results in a rare failure of $\hat{p}_\text{fail}=\num{4.1e-4}$.

\begin{figure}[t!]
    \centering
    \includesvg[inkscapelatex=false, width=\textwidth]{figures/toys/plot-combined-333-himmelblau-15-probability.svg}
    \adjustbox{trim=0 0 0 15mm, clip}{\includesvg[inkscapelatex=false, width=\textwidth]{figures/toys/plot-combined-333-truth-surrogate-himmelblau-15-probability.svg}}
    \caption{Test on a probabilistic-valued system (the \textsc{Mixture} (Himmelblau) problem).}
    \label{fig:prob_valued}
\end{figure}

\subsection{Test on Probabilistic-Valued System}\label{sec:prob_valued_sys}
As mentioned in \cref{sec:problem}, the GP construction and proposed acquisition functions were designed to estimate failure probability over binary-valued systems that indicate failure, but the same techniques are applicable when the system outputs a probabilistic value of failure (that can be interpreted as confidence in the output, distance to failure boundary, or stochasticity of the system---which has been addressed by similar approaches from \citet{gong2022sequential}).
Using the same Himmelblau function \cite{himmelblau1972applied} defined for the \textsc{Mixture} problem in \cref{sec:toys}, we change the system to output a measure of failure (where $f(\vec{x}) \ge 0.5$ means failure):  $f(\vec{x}) = \operatorname{logit}^{-1}\left(c - \left((x_1^2 + x_2 - 11)^2 + (x_1 + x_2^2 - 7)^2\right) \right)$ for the threshold $c=15$ (same as the previous problem) and passing the output through a sigmoid to interpret it as a probability with a steepness of $s=1/c$.
\Cref{fig:prob_valued} illustrates this example with a final relative error of $\hat{\Delta}_\text{fail} = \num{0.012}$, a falsification rate of $53.6\%$ of samples, an input coverage of $C_\text{input}=0.653$, and output coverage of $C_\text{output}=0.9998$.
This example is tested on the \textsc{Mixture} (Himmelblau) problem, where the output is a probability value indicating distance to the failure boundary at $f(\vec{x}) = \num{0.5}$.
The top row illustrates BSV and the FSAR acquisition functions after $N=\num{999}$ true observations (shown as red/green squares), with a uniform operational model $p$ shown as subplots.
The uniform model helps highlight that the failure region sampling is now more influenced by those failures that are farther away from the failure threshold $c$, shown as yellow peaks.
The bottom row shows the surrogate model and ground truth, where ``soft'' is the probabilistic output and ``hard'' is the binary failure classification.

\subsection{Real-World Case Study Results}

After empirically validating the BSV algorithm on the example problems with access to the ground truth, we now report the performance on a real-world example: a runway detection system.
\Cref{fig:rwd_surrogate} shows the final surrogate after running BSV for $T=333$ iterations (resulting in $999$ sampled points).
The algorithm focused the search budget on the highly likely regions of the design space and found several disjoint failure modes.
We can efficiently determine the most-likely failure, which is indicated in \cref{fig:rwd_surrogate}, and found $571$ failures out of $999$ evaluations, shown in \cref{tab:results_rwd} as the failure rate $R_\text{fail} = 57.2 \%$.
The primary goal, estimating failure probability, is shown to quickly converge in \cref{fig:rwd_p_estimates} after just over $400$ system evaluations.
The final estimated failure probability was $\hat{p}_\text{fail} = \num{5.8e-3}$.
If we instead used Monte Carlo sampling of $p$, we would expect to find only about $6$ failures in the $999$ system evaluations.

One way to characterize the spread of failures is to plot the log-likelihood of the observed failures under the operational model.
Shown in \cref{fig:ll_distribution}, the right skewed peak of the distribution indicates that the failures that were found have high likelihood and thus are more useful failures to fix first before system deployment.
Five different iterations of the BSV algorithm and acquisition functions for the RWD system are illustrated in \cref{fig:bsv_rwd}.
Red indicates failures predicted by the probabilistic surrogate model and lighter colors indicate maximums for the acquisition functions.
The red and green squares overlaid on the surrogate are the true system evaluations and the red circles in the acquisition functions show the next selected point.
Notice the low uncertainty in the concentration of points around the likely glide slope region (the operational models for glide slope and distance to runway are shown as subplots).
The likelihood decay in the boundary refinement acquisition is illustrated as the spread of the likelihood influence as it dissipates over time.
Finally, the refined failure region using $N=\num{999}$ samples represents the predicted distribution of failures.
\Cref{tab:results_rwd} reports the safety validation metrics, noting that $\hat{\Delta}_\text{fail}$ and $C_\text{output}$ are not reported since we do not have access to the true failure boundaries of the system.

\begin{table*}[t!]
\caption{Runway detection safety validation metrics using BSV.}\label{tab:results_rwd}
\vspace*{-5mm}
\begin{center}
\begin{threeparttable}
    \begin{small}
    \begin{tabular}{@{}cccc@{}}
    \toprule
        $R_\text{fail}$ & $\mathcal{L}^*$ & $\hat{p}_\text{fail}$ & $C_\text{input}$ \\
        \midrule
        $0.572$ & $0.02$ & $\num{5.8e-3}$ & $0.681$\\
        \bottomrule
    \end{tabular}
    \end{small}
\end{threeparttable}
\end{center}
\end{table*}

\begin{figure}[t!]
  \centering
  \begin{subfigure}[c]{0.49\textwidth}
    \centering
    \includesvg[inkscapelatex=false, width=\textwidth]{figures/rwd/mlf.svg}
    \caption{The final surrogate for RWD using $\num{999}$ data points.}
    \label{fig:rwd_surrogate}
  \end{subfigure}
  \hfill
  \begin{subfigure}[c]{0.49\textwidth}
    \centering
    \begin{subfigure}[t]{\textwidth}
      \centering
      \resizebox{\textwidth}{!}{%
          \input{figures/rwd/ll_distribution}
      }
      \caption{Distribution of $\log p(\vec{x})$ for observed failures.}
      \label{fig:ll_distribution}
    \end{subfigure}
    \vspace*{4mm}
    \vfill
    \begin{subfigure}[b]{\textwidth}
      \centering
      \resizebox{\textwidth}{!}{%
          \input{figures/rwd/rwd_p_estimates}
      }
      \caption{Convergence of the probability of failure estimate.}
      \label{fig:rwd_p_estimates}
    \end{subfigure}
  \end{subfigure}
  \caption{Results using Bayesian safety validation on the runway detection system.}
  \label{fig:rwd_results}
\end{figure}

\begin{figure}[pht!]
    \centering
    \setlength{\tabcolsep}{-1.5em} 
    \begin{tabular}{@{}cl@{}}
        \noindent\parbox[c]{\hsize}{\includesvg[inkscapelatex=false, width=0.92\textwidth]{figures/rwd/plot-combined-1.svg}} & {\scriptsize$\mathbf{t=1}$} \\
        \adjustbox{trim=0 0 0 15mm, clip}{\noindent\parbox[c]{\hsize}{\includesvg[inkscapelatex=false, width=0.92\textwidth]{figures/rwd/plot-combined-3.svg}}} & {\scriptsize$\mathbf{t=3}$} \\
        \adjustbox{trim=0 0 0 15mm, clip}{\noindent\parbox[c]{\hsize}{\includesvg[inkscapelatex=false, width=0.92\textwidth]{figures/rwd/plot-combined-9.svg}}} & {\scriptsize$\mathbf{t=9}$} \\
        \adjustbox{trim=0 0 0 15mm, clip}{\noindent\parbox[c]{\hsize}{\includesvg[inkscapelatex=false, width=0.92\textwidth]{figures/rwd/plot-combined-100.svg}}} & {\scriptsize$\mathbf{t=100}$} \\
        \adjustbox{trim=0 0 0 15mm, clip}{\noindent\parbox[c]{\hsize}{\includesvg[inkscapelatex=false, width=0.92\textwidth]{figures/rwd/plot-combined-333.svg}}} & {\scriptsize$\mathbf{t=333}$} \\
    \end{tabular}
    \caption{Bayesian safety validation applied to the runway detection problem.}
    \label{fig:bsv_rwd}
\end{figure}
\setlength{\tabcolsep}{6pt} 

Results show that we can characterize failure regions, generate a set of likely failures, compute the most-likely failure, and use the surrogate model to estimate the probability of failure of the runway detector in a small number of samples; only use $999$ samples in our experiments.
Post-analysis could even further characterize the failure boundary by focusing on the likely region centered around the glide slope angle of $3$ degrees.
We demonstrate the BSV algorithm on a two-dimension case but this work could be scaled to higher-dimensional problems that incorporate additional environmental parameter models such as roll angle, time-of-day, weather, and across different airport runways.
\citet{binois2022survey} provide a survey on methods, challenges, and guidelines when modeling high-dimensional problems using Gaussian processes for Bayesian optimization.

\section{Conclusion}\label{sec:conclusion}
In this paper, we frame the black-box safety validation problem as a Bayesian optimization problem and introduce Bayesian safety validation (BSV) to build a probabilistic surrogate model that predicts system failures and uses importance sampling to efficiently estimate the failure probability.
In the process, we propose a set of acquisition functions, called \textit{failure search and refinement} (FSAR), that each help achieve the safety validation tasks by covering the design space to search for failures and refining likely failure boundaries and regions.
The Gaussian process construction allows us to analytically derive the predicted failure boundaries and we show that the combination of acquisition functions is important to find more failures, find more likely failures, and minimize error in the failure probability estimate.
Primarily interested in cases where the black-box system only outputs a binary indication of failure, we also show that our method works well in the less restrictive case where the system outputs a real-valued measure of failure confidence, severity, or distance.
BSV is also applicable when system failures are stochastic, thus the idea of deterministic ``failure regions'' becomes irrelevant.
In the case of stochastic system failures, self-normalizing importance sampling is used to compute the proposal weights, which scales better to higher dimensional problems.
This technique was applied to validate an image-based neural network runway detection system in simulation.
Alongside traditional DO-178C procedures \cite{do178c}, this work is currently being used to supplement the FAA certification process of an autonomous cargo aircraft \cite{durand2023formal}.

The use of a simulator allows us to quickly assess the performance of systems, yet validating that the simulator correctly captures reality is an important research challenge being address by the sim-to-real community \cite{zhao2020sim}.
For the runway detection case, one way to perform this validation would be to run BSV and select a representative subset of safe-to-fly points to flight test, then compare their outputs.
We emphasize that the exact values of the most-likely failure likelihood and the estimated probability of failure are largely dependent on the choice of operational model and learning these models from collected flight data would provide a more realistic understanding of the true probability of failure.
This work is open sourced and available at {\small\url{https://github.com/sisl/BayesianSafetyValidation.jl}}.

\section*{Appendix}

\subsection{Open-Source Interface}
This work has been open sourced\footnote{The package and experiment code are available at \url{https://github.com/sisl/BayesianSafetyValidation.jl}.} as a Julia package\footnote{We make use of the \texttt{GaussianProcesses.jl} package \cite{fairbrother2022gaussianprocesses} and the \texttt{AbstractGPs.jl} package \cite{widmann2023abstractgps}.} to be applied to other of black-box systems and is intended to fit into the suite of safety validation tools when considering autonomous aircraft certification.
To extend to another system, a user can implement the following interface:

\begin{juliaframe}
\begin{lstlisting}[language=JuliaLocal, style=julia]
abstract type SystemParameters end
function reset() end
function initialize() end
function generate_input(sample::Vector)::Input end
function evaluate(input::Input)::Vector{Bool} end
\end{lstlisting}
\end{juliaframe}
Below is an example of setting up a system with a two-dimensional operational model, running the Bayesian safety validation algorithm to learn the surrogate, and then computing the three safety validation tasks.
\begin{juliaframe}
\begin{lstlisting}[language=JuliaLocal, style=julia]
using BayesianSafetyValidation
system_params = RunwayDetectionSystemParameters() # defined by user as <: SystemParameters
model = [OperationalParameters("distance", [0.1, 4], TruncatedNormal(0, 1.0, 0, 4)),
         OperationalParameters("slope", [1, 7], Normal(3, 0.5))]
surrogate  = bayesian_safety_validation(system_params, model; T=100)
X_failures = falsification(surrogate.x, surrogate.y)
ml_failure = most_likely_failure(surrogate.x, surrogate.y, model)
p_failure  = p_estimate(surrogate, model)
\end{lstlisting}
\end{juliaframe}

\section*{Acknowledgments}
We would like to thank Jean-Guillaume Durand and Anthony Corso for their thoughtful and continuous insights into this research.
We thank Rob Timpe and Alexander Bridi for the development of the runway detection neural network.
We also thank Harrison Delecki for inspiration of \cref{fig:safety_validation}.
This work was supported by funding from Xwing, Inc.

\bibliography{references}

\end{singlespacing}

\end{document}